\documentclass{article} 
\usepackage[OT1]{fontenc}
\usepackage{iclr2027_conference,times}

\usepackage{amsmath,amsfonts,bm}

\def\eqref#1{equation~\ref{#1}}

\def\1{\bm{1}}

\DeclareMathAlphabet{\mathsfit}{\encodingdefault}{\sfdefault}{m}{sl}
\SetMathAlphabet{\mathsfit}{bold}{\encodingdefault}{\sfdefault}{bx}{n}

\usepackage{hyperref}
\usepackage{url}
\usepackage{booktabs}
\usepackage{graphicx}
\usepackage{array}
\usepackage{amssymb}
\usepackage{float}
\usepackage[most]{tcolorbox}
\usepackage{multirow}

\title{VibeEdit: Image Editing with Canvas Instructions}

\author{%
\parbox{\textwidth}{%
\raggedright
Jinjing Zhao$^{1}$\thanks{Equal contribution.}\quad
Fangyun Wei$^{2}$\footnotemark[1]\quad
Yitong Wang$^{3}$\quad
Xiuyu Wu$^{4}$\quad
Yunuo Chen$^{5}$\quad
Yang Yue$^{6}$\quad
Sirui Zhang$^{7}$\quad
Wenbo Wang$^{1}$\quad
Hongyang Zhang$^{8}$\quad
Dong Chen$^{2}$\quad
Yan Lu$^{2}$\quad
Chang Xu$^{1}$\thanks{Corresponding author.}\\[0.7em]
\normalfont
$^{1}$University of Sydney\quad
$^{2}$Microsoft Research\quad
$^{3}$Fudan University\\[0.2em]
$^{4}$Nankai University\quad
$^{5}$Shanghai Jiao Tong University\quad
$^{6}$Tsinghua University\\[0.2em]
$^{7}$University of Science and Technology of China\quad
$^{8}$University of Waterloo\\[0.7em]
{\footnotesize
\mbox{\texttt{\{jzha0100,wwan0412,c.xu\}@sydney.edu.au}}\quad
\mbox{\texttt{\{fawe,doch,yanlu\}@microsoft.com}}\quad
\mbox{\texttt{wangyitong23@m.fudan.edu.cn}}\quad
\mbox{\texttt{xiuyuwu@mail.nankai.edu.cn}}\quad
\mbox{\texttt{cyril-chenyn@sjtu.edu.cn}}\quad
\mbox{\texttt{yueyang22@mails.tsinghua.edu.cn}}\quad
\mbox{\texttt{zsr200901@mail.ustc.edu.cn}}\quad
\mbox{\texttt{hongyang.zhang@uwaterloo.ca}}
}
\\[0.7em]
\makebox[\linewidth][c]{%
  {\small
  \textbf{Project page:}\;
  \href{https://zhaojingjing713.github.io/VibeEdit/}
       {\texttt{https://zhaojingjing713.github.io/VibeEdit/}}%
  }%
}
}
}

\iclrfinalcopy 
\begin{document}

\maketitle
\fancyhead{}
\renewcommand{\headrulewidth}{0pt}

\vspace{-2.0em}
\begin{figure}[!htbp]
    \centering
    \includegraphics[width=0.99\textwidth]{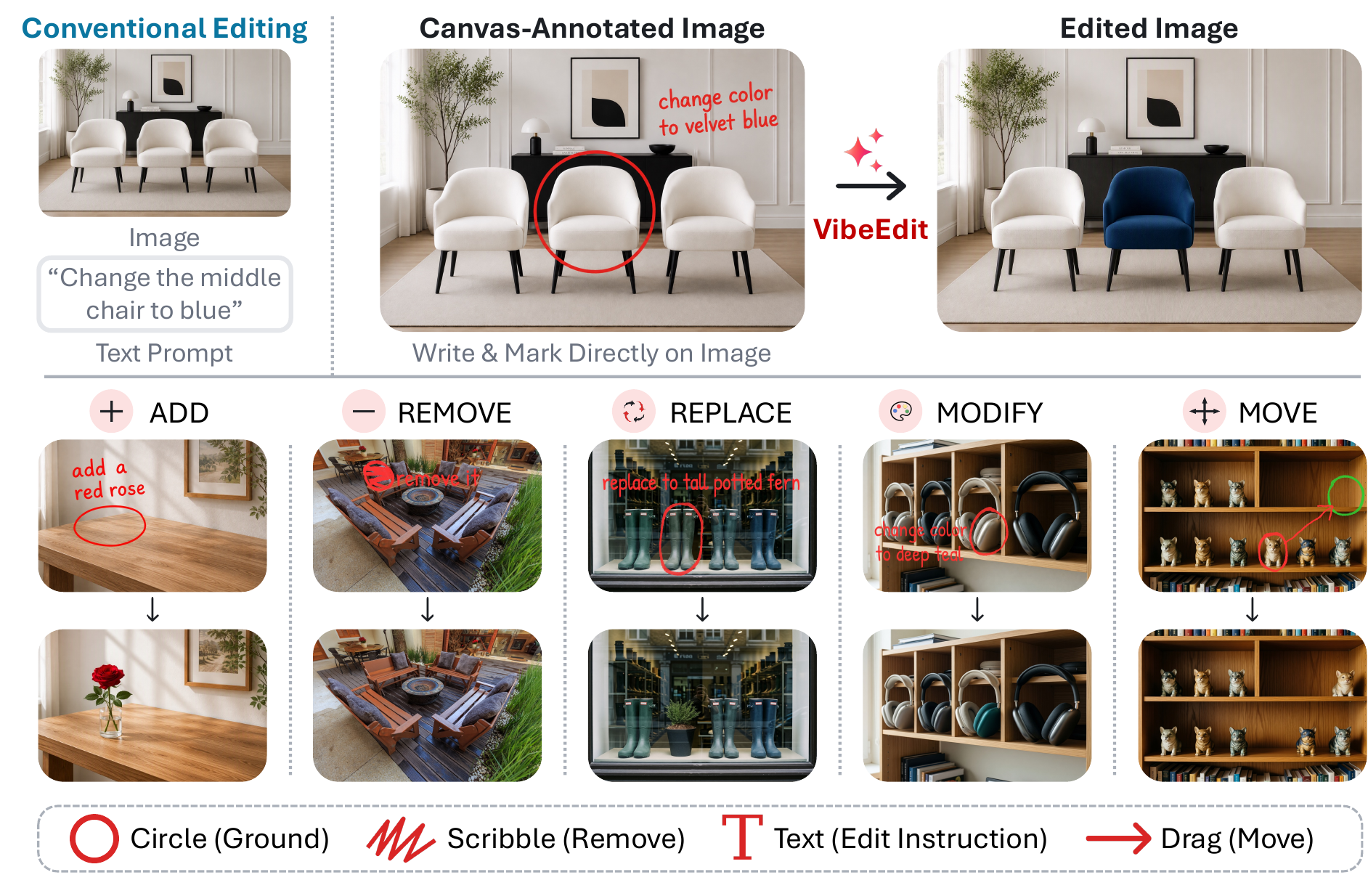}
    \vspace{-4mm}
    \caption{\textbf{VibeEdit} enables image editing without requiring a separate text prompt. Users express editing intent directly on the image through \underline{canvas instructions}, including \textit{spatial marks} such as circles, scribbles, and drag gestures, optionally accompanied by short \textit{text notes}. Given the resulting \textit{canvas-annotated image}, VibeEdit supports diverse editing operations, including object addition, removal, replacement, attribute modification, and movement. More visualizations are provided in Figure~\ref{fig:more-demo} in the Appendix.} 
    \label{fig:teaser}
\end{figure}
\begin{abstract}
In text-guided image editing, describing the desired change is often straightforward, but identifying the intended object or region can be cumbersome, especially when several objects look alike. We introduce a new image editing interface that lets users place spatial marks and optional short notes directly on the image. Together, these annotations form a \emph{canvas instruction} that specifies where to edit and what to change. Our editor, \emph{VibeEdit}, follows these instructions to perform object addition, removal, replacement, attribute modification, and movement without a separate text prompt. We construct 1.55 million source--target edit pairs with object masks and structured edit descriptions, from which we render canvas instructions during training. We adapt Qwen-Image-Edit with layer-decoupled conditioning that separately encodes source images and canvas instructions for image editing. We train the model with region-weighted supervised fine-tuning, followed by rubric-guided reinforcement learning to improve edit completion, local edit quality, and preservation of unedited regions. We evaluate VibeEdit on an independently constructed, human-curated benchmark of 419 cases emphasizing target selection among similar objects. VibeEdit achieves a VLM rubric score of 79.9 and an outside-region PSNR of 32.8 dB, compared with 67.4 and 24.0 dB for FireRed, the highest-scoring text-instructed baseline in our evaluation.
\end{abstract}
\section{Introduction}
\label{sec:introduction}

Instruction-based image editors translate natural-language requests into image changes \citep{brooks2023instructpix2pix,wu2025qwenimage,longcat2025image,blackforestlabs2026flux2klein,qiao2026firered,song2026joyai,liu2025step1x,openai2025gptimage}. For a localized edit, users must specify both what to change and where to apply it. Describing the desired change is often straightforward, but identifying the intended object can require detailed descriptions of its position and surroundings, especially when several objects look alike. Points and brush strokes provide direct spatial control \citep{pan2023draggan,liu2025magicquill}, while language expresses the desired change. Combining these complementary cues directly on the image offers a natural way to link an edit to its target.

We introduce a new image editing interface that lets users place spatial marks and optional short notes directly on the image (Figure~\ref{fig:teaser}). Together, these annotations form a \emph{canvas instruction} that specifies where to edit and what to change. Circles identify objects or regions, scribbles indicate removal, and arrows specify movement. Short notes placed near the marks describe the desired content or attributes. For example, a user can circle a chair and write a new color beside it, or mark an empty region and name an object to add. These marks need not follow precise object boundaries. To support this interface, we develop \emph{VibeEdit}, an image editor that follows canvas instructions to perform object addition, removal, replacement, attribute modification, and movement without a separate text prompt.

Training VibeEdit requires paired examples that connect canvas instructions to the intended edits. We construct a corpus of 1.55 million source--target edit pairs by selecting editable objects and generating edited targets with existing models. Each pair includes object masks and structured edit descriptions. During training, we use these records to render spatial marks and short notes, sampling either editing direction where applicable. We vary stroke shapes, widths, and typefaces, including handwriting and print styles, to expose the model to different annotations for the same edit.

Following canvas instructions requires the editor to interpret annotations in context without reproducing them in the output. We adapt Qwen-Image-Edit \citep{wu2025qwenimage} with layer-decoupled conditioning: the vision--language encoder interprets the annotated image, while the generative pathway separately encodes the source image and canvas annotations. This design provides spatial guidance while retaining a clean source reference. Because most pixels remain unchanged in a localized edit, we use region-weighted supervised fine-tuning to emphasize the edit region and annotation removal while retaining supervision over the full image. A visually plausible result can still target the wrong object or alter unrelated content. We therefore refine the editor with rubric-guided reinforcement learning \citep{feng2025rubricrl,zheng2025diffusionnft}, using rewards for edit completion, local edit quality, and preservation of unedited regions.

We evaluate VibeEdit on an independently constructed, human-curated benchmark of 419 cases. The benchmark emphasizes scenes with similar objects to test whether the editor applies the requested change to the intended instance. We jointly assess edit success, local edit quality, and preservation of unedited regions, since preserving the source alone does not establish that an edit was completed. VibeEdit achieves an overall VLM rubric score of 79.9 and an
outside-region PSNR of 32.8 dB, outperforming all evaluated
baselines on both metrics, even when baselines receive detailed
external text instructions specifying the edit target and desired
change.

Our contributions are threefold:
\begin{itemize}
    \item We introduce a new image editing interface in which spatial marks and optional short notes form canvas instructions, specifying edit targets and desired changes directly on the image across five editing operations. This interface is complementary to conventional text prompting, providing direct spatial grounding when language alone is cumbersome or ambiguous.
    \item We construct 1.55 million source--target edit pairs for canvas instruction training and develop VibeEdit by adapting Qwen-Image-Edit with layer-decoupled conditioning, region-weighted supervised fine-tuning, and rubric-guided reinforcement learning.
    \item We construct an independent, human-curated benchmark of 419 cases emphasizing target selection among similar objects. VibeEdit outperforms the evaluated baselines in both overall VLM rubric score and outside-region PSNR.
\end{itemize}
\section{Related Work}
\label{sec:related-work}

\noindent\textbf{Instruction-based image editing.}
Large-scale text-to-image pretraining supplies the generative prior for modern
editors. Recent systems span publicly released HunyuanImage~3.0, Z-Image, and
Lens, alongside proprietary Qwen-Image-2.0 and Seedream~4.0, exploring
autoregressive modeling, multimodal diffusion, and unified generation and editing
\citep{cao2025hunyuanimage,zimage2025,chen2026lens,zhao2026qwenimage20,
seedream2025seedream40}.
Diffusion editors translate language instructions into image transformations.
InstructPix2Pix learns this mapping from synthetic instruction pairs
\citep{brooks2023instructpix2pix}, while MagicBrush
\citep{zhang2023magicbrush} and UltraEdit \citep{zhao2024ultraedit} broaden
instruction-editing datasets. Recent foundation editors further improve
instruction following and source-image fidelity
\citep{wu2025qwenimage,longcat2025image,blackforestlabs2026flux2klein,
qiao2026firered,song2026joyai,liu2025step1x,openai2025gptimage,
google2025nanobanana}. Text-only prompts must describe both the desired change
and its referent. VibeEdit instead lets users place short text notes directly
beside the intended edit targets.

\noindent\textbf{Spatial control and visual instructions.}
Spatial interfaces generally follow three patterns. First, operation-specific
methods encode geometric correspondence through handle and target points.
DragGAN \citep{pan2023draggan} and DragDiffusion
\citep{shi2023dragdiffusion} focus on object movement and deformation. Second,
region-guided visual-instruction methods combine image-space cues with separate
text. Inter-Edit
\citep{liu2026interedit}, FineEdit \citep{xu2026fineedit}, and scribble-guided
editing \citep{xu2026scribble} use masks, boxes, or scribbles for localization,
while VIBE \citep{zhang2026vibe} evaluates broader visual cues with textual task
instructions. Third, visual-interaction systems structure edit intent through
graphical controls. The MagicQuill series
\citep{liu2025magicquill,liu2026magicquillv2} explores richer interactions: the
original system combines brush modes with an inferred editing prompt, while its successor couples language with layered content, spatial, and color
controls. By contrast, VibeEdit combines circles, scribbles, and movement arrows with open-vocabulary
handwritten instructions. Rather than using image-space cues only for
localization, it also places semantic guidance directly on the canvas across
five edit types.

\noindent\textbf{Reinforcement learning for diffusion models.}
Reward-based post-training aligns diffusion and flow models with visual
objectives difficult to express through likelihood training. DDPO
\citep{black2023ddpo} formulates denoising as a sequential decision process.
Flow-GRPO and DanceGRPO adapt group-relative policy optimization to diffusion
and flow-based generation \citep{liu2025flowgrpo,xue2025dancegrpo}.
OP-GRPO \citep{zhang2026opgrpo} improves sample reuse through off-policy
trajectory replay. DiffusionNFT \citep{zheng2025diffusionnft} instead optimizes
a forward-process objective using positive and implicit negative predictions,
without storing full denoising trajectories. For reward design, RubricRL
\citep{feng2025rubricrl} decomposes prompt requirements into criteria evaluated
independently by a multimodal judge. We adapt forward-process updates and
decomposed rewards to localized editing, where edit completion must be
balanced with source preservation.

\section{Data and Benchmark Construction}
\label{sec:dataset}

We construct three datasets for VibeEdit across five edit types: a supervised learning corpus for training the editor to follow canvas instructions, an RL set selected by reward variation, and a human-curated benchmark to evaluate instruction following and target selection among similar objects.

\begin{figure}[!t]
    \centering
    \vspace{-1mm}
    \includegraphics[width=1.0\textwidth]{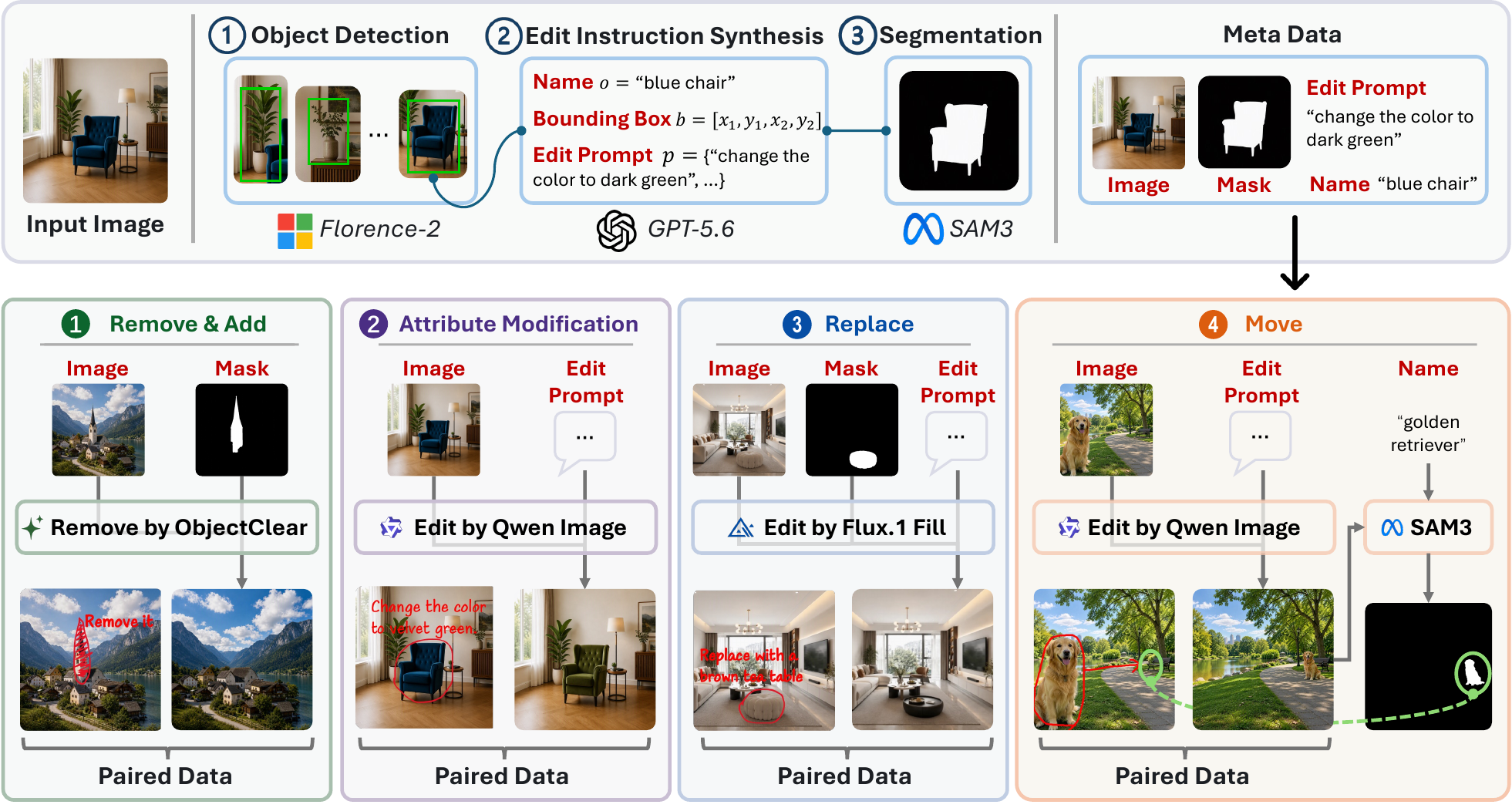}
    \vspace{-7mm}
    \caption{Supervised learning edit-pair construction. A shared pipeline performs object
    selection, edit prompt synthesis, and segmentation. Operation-specific pipelines then generate source--target pairs for addition, removal, attribute modification, replacement, and movement. Canvas annotations are shown for illustration and are rendered online during training.}
    \vspace{-8mm}
    \label{fig:data}
\end{figure}

\subsection{Supervised Learning Data}
\label{sec:edit-pair-construction}

We synthesize edit pairs offline and render canvas instructions online
during training. Figure~\ref{fig:data} shows the overall data construction pipeline. Each stored record contains a source image $I$, an edited
target $I_{\mathrm{edit}}$, object masks, and structured edit descriptions.
This representation separates the underlying edit from its annotation
style, allowing the same pair to support varied canvas instructions.
Where applicable, we also retain the information needed to render
instructions in both editing directions.

\noindent\textbf{Object selection and prompt synthesis.}
In Figure~\ref{fig:data}, Florence-2 \citep{xiao2024florence2}
generates open-vocabulary object proposals. We discard excessively small
or large proposals, assign identifiers to the remaining candidates, and provide the indexed image and optional candidate crops to GPT-5.6-sol~\citep{openai2026gpt56}. The model selects an object that is recognizable,
sufficiently visible, mostly within the frame, and spatially separable
from its surroundings. It records the object's name $o$ and, when needed for edit synthesis, generates an edit prompt $p$. SAM~3 \citep{carion2025sam3} produces an object mask $M$ from the selected box.
Detailed selection rules are provided in Appendix~\ref{sec:selection-prompts}.

\noindent\textbf{Addition and removal.}
We select objects with clean boundaries and backgrounds suitable for
reconstruction, avoiding tightly entangled instances and prominent
attached shadows or reflections. ObjectClear \citep{zhao2026objectclear}
inpaints the masked region of $I$ to produce an image without the selected
object. The original-to-inpainted direction provides removal supervision;
reversing the pair provides addition supervision. Each pair thus
supports both operations, with the added object and its context drawn
from the same original image. We retain the object's name and visual
description.

\noindent\textbf{Attribute modification.}
GPT-5.6-sol selects one visible attribute from color, material,
texture, pattern, state, or object-local style, and specifies a distinct,
verifiable target value. The edit prompt identifies the object and the
requested change while instructing the editor to preserve its identity,
geometry, pose, and position. Qwen-Image-Edit \citep{wu2025qwenimage}
applies the prompt $p$ to $I$ without using $M$ as a generation constraint.
We retain $M$ for localization and store both the original and target
attribute values.

\noindent\textbf{Object replacement.}
GPT-5.6-sol selects a contextually plausible replacement from a
different category and at a scale appropriate for the scene. FLUX.1 Fill
\citep{blackforestlabs2024fluxfill} generates the replacement described
by the edit prompt within an expanded version of $M$, preserving the
surrounding scene. We retain descriptions of both the original and
replacement objects.

\noindent\textbf{Object movement.}
GPT-5.6-sol selects an object that can move as a rigid unit and a
clearly separated destination on a suitable, unoccupied support surface.
Qwen-Image-Edit~\citep{wu2025qwenimage} follows the edit prompt to relocate the object and produce
$I_{\mathrm{edit}}$. The original object mask serves as the source mask
$M_{\mathrm{src}}$. To locate the object after editing, GPT-5.6-sol
provides a description of its stable appearance attributes and intended
destination. We use this description to prompt SAM~3~\citep{carion2025sam3} on
$I_{\mathrm{edit}}$, obtaining the destination mask $M_{\mathrm{tgt}}$.
The two masks specify the endpoints of the movement annotation;
reversing the edit swaps their roles.

\noindent\textbf{Quality control.}
We use GPT-5.6-sol to assess the synthesized edit pairs and filter
out low-quality examples. Since filtering may not eliminate all
errors, we complement this data-level screening with rubric-guided
RL after supervised training. By favoring outputs with successful
edits, high local quality, and preserved surrounding content,
RL steers the model toward the desired editing behavior and
helps mitigate the influence of residual training noise.

\noindent\textbf{Canvas instruction rendering.}
During training, we sample a valid edit direction and use the corresponding
masks and edit descriptions to render a canvas instruction $C$. Circles
and short notes specify addition, attribute modification, and replacement.
A circled scribble accompanied by a short deletion note indicates removal.
Two circles connected by a directed arrow specify movement.
We vary stroke shapes, widths, and typefaces during rendering.

We form the canvas-annotated image $A=I\oplus C$, where $\oplus$
denotes rendering the spatial marks and text notes in $C$ onto
the source image $I$. These annotations specify the edit without
a separate text prompt. We retain $I$ and $C$ separately for
the layer-decoupled conditioning in Section~\ref{sec:method-conditioning}.
The final corpus contains 1,554,062 stored image pairs.
Appendix~\ref{sec:supp-data} details the instruction grammar,
rendering procedure, and corpus statistics.

\subsection{Reinforcement Learning Data and Benchmark}
\label{sec:rl-data}

\noindent\textbf{RL data selection by reward variation.}
We select RL conditions based on reward variation across edits
sampled from our base model trained with supervised learning. For each candidate image--canvas instruction pair
$c=(I,C)$, we generate $K$ offline rollouts, score them with the structured
reward defined in Section~\ref{sec:method-rl}, and compute
$\sigma(c)=\operatorname{Std}_{k=1}^{K}[R(\hat I_k;c)]$.
Within each edit type, we retain the candidates with the highest
$\sigma(c)$. Selection thus favors reward differences among rollouts
of the same condition, which provide the relative training signal for
DiffusionNFT, rather than low mean rewards. The resulting RL set contains
3,520 conditions, with 704 for each edit type.

\noindent\textbf{Human-curated VibeEdit benchmark.}
We construct a 419-case benchmark independently of the training
corpus, using images manually collected and selected from online sources. The benchmark emphasizes scenes containing multiple objects
from the same category or with similar appearances. 
For each case, human annotators specify the edit region with a
mask and describe the intended change. We use these annotations
to render spatial marks and short notes as needed, forming the
canvas instruction. For comparison with text-instructed baselines,
we additionally prepare a detailed external text instruction for
each case, specifying the target object's location within the scene
and the requested edit. These baselines receive the source image
and external text instruction, while VibeEdit receives the source
image and canvas instruction without a separate text prompt.
The benchmark tests whether an editor interprets
the requested change and applies it to the intended object or region.
Per-task counts and external text-instruction lengths are reported in
Appendix~\ref{sec:data-statistics}.

\section{Method}
\label{sec:method}

\begin{figure}[t]
    \centering
    \vspace{-2mm}
    \includegraphics[width=1.0\textwidth]{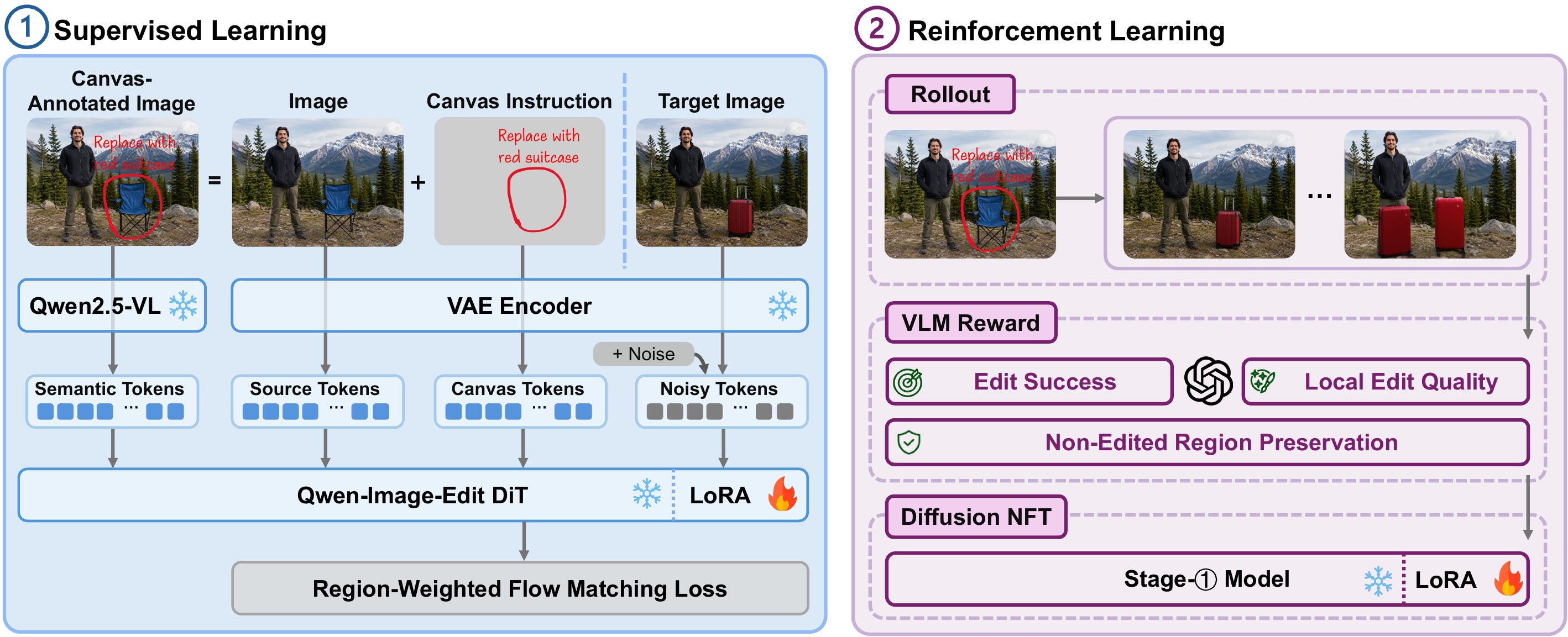}
    \vspace{-6mm}
    \caption{VibeEdit training pipeline. \textbf{Left:} the vision--language
    encoder processes annotated image $A$, while the VAE separately
    encodes source image $I$ and rasterized instruction $\bar C$.
    The target image supervises region-weighted training.
    \textbf{Right:} the base model is refined with DiffusionNFT using
    rewards for edit success, outside preservation, and local edit quality.}
    \vspace{-5mm}
    \label{fig:pipe}
\end{figure}

Given a source image $I$ and a canvas instruction $C$, VibeEdit produces
an edited image $\hat I=f_\theta(I,C)$ without a separate text prompt.
We denote the canvas-annotated image by $A=I\oplus C$.
As shown in Figure~\ref{fig:pipe}, layer-decoupled conditioning
uses the vision--language encoder to interpret the annotated image
and the VAE to encode the source image and canvas instruction
separately for generation. We then train the model in two stages:
region-weighted supervised learning emphasizes the requested local
changes, and rubric-guided RL further refines the model using
rewards for edit success, local edit quality, and preservation
of unedited regions.

\subsection{Layer-Decoupled Conditioning}
\label{sec:method-conditioning}

Canvas annotations must guide the edit without appearing in the output.
Encoding only the canvas-annotated image $A$ in the generative pathway mixes
source content with instruction marks. We therefore use the canvas-annotated
image for semantic interpretation and separate the image from
the annotations for generative conditioning.

The frozen Qwen2.5-VL~\citep{bai2025qwen25vltechnicalreport} encoder processes $A$ to produce semantic tokens
$h=E_{\mathrm{VL}}(A)$. For the generative pathway, we rasterize $C$
on a neutral gray background to obtain $\bar C$, then encode the two
layers separately as $z_I=E_{\mathrm{VAE}}(I)$ and
$z_C=E_{\mathrm{VAE}}(\bar C)$. This provides a clean source reference
and an explicit representation of the annotation layout, while the
semantic tokens convey the instruction in the context of the scene.

Following the conditioning scheme of Qwen-Image-Edit~\citep{wu2025qwenimage}, the DiT receives noisy
target tokens, source tokens $z_I$, canvas tokens $z_C$, and semantic
tokens $h$. The noisy target tokens use timestep $t$, while source
and canvas tokens use timestep zero. We freeze the VAE, the vision--language encoder, and the DiT weights, and train rank-128 LoRA~\citep{hu2021lora} adapters on the
DiT attention projections and modulation layers.

\subsection{Region-Weighted Supervised Fine-Tuning}
\label{sec:method-sft}

For a localized edit, most target pixels remain unchanged from the
source. A spatially uniform loss can therefore underemphasize the
requested change. We assign greater weight to the edited object and
rendered annotations to emphasize edit completion and annotation
removal, while retaining supervision over the full image.

Let $z_0=E_{\mathrm{VAE}}(I_{\mathrm{edit}})$ be the target latent.
For a sampled timestep $t$ with noise level $\sigma_t$ and Gaussian
noise $\epsilon$, we define the noisy latent and target velocity as
\begin{equation}
    z_t=(1-\sigma_t)z_0+\sigma_t\epsilon,
    \qquad v^*=\epsilon-z_0.
\end{equation}
We form a binary supervision mask
$M_{\mathrm{sup}}=M_{\mathrm{obj}}\lor M_{\mathrm{ann}}$,
where $M_{\mathrm{obj}}$ covers the edited object and
$M_{\mathrm{ann}}$ covers all rendered strokes and text.
For movement, $M_{\mathrm{obj}}$ is the union of the source and
destination object masks. We dilate $M_{\mathrm{sup}}$ and map it
to latent resolution to obtain $\widetilde M$. Dilation extends
the upweighted region around object boundaries and annotations to
accommodate approximate marks, nearby edit effects, and VAE
downsampling. The supervised learning objective is
\begin{equation}
\mathcal{L}_{\mathrm{SFT}}
=\mathbb{E}_{t,\epsilon}
\left[
    \left\|
    \sqrt{\mathbf{1}+\lambda\widetilde M}\odot
    \left(v_\theta(z_t;h,z_I,z_C,t)-v^*\right)
    \right\|_2^2
\right],
\label{eq:mask-flow-loss}
\end{equation}
where $\widetilde M$ is broadcast over latent channels. The loss retains
unit weight outside the mask and increases the weight within it.
We set $\lambda=1.5$ and dilate the mask by $50$ pixels before
downsampling. The supervision mask only weights the training loss
and is not provided to the model at inference.

\subsection{Rubric-Guided Diffusion Reinforcement Learning}
\label{sec:method-rl}

Supervised learning teaches the model to follow canvas instructions.
We then refine its editing behavior through rubric-guided RL. For each condition in the RL set from Section~\ref{sec:rl-data}, the sampling policy $\pi_{\mathrm{old}}$
generates multiple edits, and a  VLM judge assigns rubric rewards.
The rollout latents are then re-noised for DiffusionNFT optimization
\citep{zheng2025diffusionnft}.

\noindent\textbf{Structured rubric reward.}
For each image--canvas instruction pair $(I,C)$ of edit type $\tau$,
$\pi_{\mathrm{old}}$ generates $K$ rollouts
$\{\hat I_k\}_{k=1}^K$. To evaluate each rollout, the VLM judge receives
the source and edited images, aligned before/after crops of the edit
region, the edit type, and the structured target description.
Movement examples include crops of both the source and destination
regions. The structured description is used only for reward computation
and is not provided to the editing policy.

Following the rubric rewards design \citep{feng2025rubricrl},
we assess each output with binary questions in three groups:
\begin{itemize}
    \item \emph{Edit success} checks whether the requested operation
    is completed at the intended location, using operation-specific
    criteria such as object presence, absence, or identity preservation.
    \item \emph{Outside preservation} checks whether unrelated objects
    and scene content remain unchanged, with no unintended edits
    or duplicates.
    \item \emph{Local edit quality} checks whether the edited region
    is visually well-formed, naturally integrated, and aesthetically
    consistent with the scene.
\end{itemize}
The judge prompt and rubric questions are provided in Appendices~\ref{sec:rubric-prompt} and~\ref{sec:complete-rubric-questions}.
Let $\mathcal{Q}$ denote the questions applicable to the current
example and $q_{k,j}\in\{0,1\}$ the judge's decision for question $j$
on rollout $k$. We average these decisions and apply a penalty for
insufficient preservation outside the edit region:
\begin{equation}
    R_k^{\mathrm{rub}}=\frac{1}{|\mathcal{Q}|}\sum_{j\in\mathcal{Q}}q_{k,j},
    \qquad
    R_k=R_k^{\mathrm{rub}}-\gamma_{\mathrm{p}}
    \mathbf{1}[\operatorname{PSNR}_{\mathrm{out}}<\tau_{\mathrm{p}}].
\label{eq:rubric-reward}
\end{equation}
Here, $\operatorname{PSNR}_{\mathrm{out}}$ compares the source image
with rollout $k$ outside a padded edit region. For movement, the
excluded region covers both the source and destination. This penalty
accounts for low-level image changes that binary VLM judgments may
miss. We set $\gamma_{\mathrm{p}}=0.3$ and
$\tau_{\mathrm{p}}=25$ dB.

\noindent\textbf{Reward normalization.}
We center rewards within each condition and normalize them by the
reward standard deviation of the corresponding edit type:
\begin{equation}
    a_k=\frac{R_k-\frac{1}{K}\sum_{i=1}^K R_i}
    {s_\tau+\delta},
    \qquad
    \rho_k=\frac{1}{2}+\frac{1}{2}
    \operatorname{clip}\!\left(\frac{a_k}{a_{\max}},-1,1\right),
\label{eq:rl-optimality}
\end{equation}
where $s_\tau$ is computed over the current rollouts of edit type
$\tau$ across workers, $a_{\max}$ is the advantage clipping threshold,
and $\delta$ ensures numerical stability. Centering compares each
rollout with others generated for the same condition, while type-wise
normalization adjusts for differences in reward scale across edit
types. Unlike the per-condition statistic $\sigma(c)$ used for offline
data selection, $s_\tau$ is computed online across rollouts of the
same edit type.

The resulting $\rho_k\in[0,1]$ controls the relative weights of the
positive and negative branches in the DiffusionNFT objective:
higher-reward rollouts receive greater weight in the positive branch.
The policy-update objective and implementation details are provided
in Appendix~\ref{sec:nft-update}.

\section{Experiments}
\label{sec:experiments}

\subsection{Experimental Setup}
\label{sec:experimental-setup}

\noindent\textbf{Training.}
We initialize VibeEdit from Qwen-Image-Edit~\citep{wu2025qwenimage} and train rank-128
LoRA~\citep{hu2021lora} adapters on the DiT attention projections and modulation layers.
Supervised learning uses aspect-ratio buckets of approximately $1024^2$ pixels,
a global batch size of 128, and a constant learning rate of
$5\times10^{-5}$ after 500 warmup steps. RL starts from the
12,000-step supervised fine-tuned checkpoint and trains a new rank-128 LoRA adapter
at $512^2$ resolution with the same learning rate. Each condition
uses 12 rollouts with eight denoising steps and classifier-free
guidance (CFG) scale 1, scored by the fixed GPT-5.4-mini~\citep{openai2026gpt54} judge.
We evaluate RL at 220 steps. Both stages use 32 NVIDIA A100 GPUs.

\noindent\textbf{Inference.}
All evaluations use $H\times W\approx1024^2$ pixels and one output
per case. Base models use 30 denoising steps and CFG scale 4;
RL models use 16 steps and CFG scale 1. Public baselines
otherwise use their released pipelines and recommended settings.

\noindent\textbf{Metrics.}
On the 419-case benchmark, GPT-5.6-sol~\citep{openai2026gpt56} serves as a fixed judge.
It answers binary rubric questions on edit success, outside
preservation, and local edit quality using source and edited
images with aligned crops. The overall VLM rubric score (\%)
averages per-case pass rates within each edit type, then equally
across the five types. Appendices~\ref{sec:rubric-prompt}
and~\ref{sec:complete-rubric-questions} detail the protocol and rubric.
We also report source--output PSNR outside a padded edit region.

\begin{table}[t]
\vspace{-4mm}
\caption{Main comparison on the VibeEdit benchmark. We tailor the input format to different model types. Visual-only baselines
receive the canvas-annotated image $A$; text-instructed baselines receive the
source image $I$ and a text instruction. VibeEdit uses $(I,C)$ without a separate
text prompt. Entries report \textbf{V} (VLM rubric score (\%)) / \textbf{P} (outside-region PSNR (dB)).
\textbf{Bold} and \underline{underlined} values indicate the best and second-best results.}
\vspace{2mm}
\label{tab:main-results}
\centering
\small
\setlength{\tabcolsep}{3.pt}
\begin{tabular}{lcccccc}
\toprule
\textbf{Method} & \textbf{Add V/P} & \textbf{Remove V/P} &
\textbf{Attribute V/P} & \textbf{Replace V/P} &
\textbf{Move V/P} & \textbf{Overall V/P} \\
\midrule
\multicolumn{7}{l}{\textit{Visual-only baselines}} \\
Qwen-Image-Edit-2511 & 17.7 / 17.0 & 22.1 / 14.3 & 28.1 / 17.4 & 21.1 / 16.5 & 10.0 / 16.1 & 19.8 / 16.3 \\
LongCat-Image-Edit & 8.8 / 15.0 & 39.0 / 18.9 & 29.5 / 14.8 & 27.8 / 16.2 & 10.5 / 13.5 & 23.1 / 15.6 \\
FLUX.2-klein-9B & 30.0 / 24.1 & 34.3 / 27.6 & 56.9 / \underline{26.8} & 43.3 / 25.7 & 23.8 / \textbf{27.9} & 37.6 / 26.4 \\
FireRed-Image-Edit-1.0 & 26.9 / 17.9 & 35.4 / 17.1 & 28.6 / 13.4 & 19.0 / 16.6 & 4.2 / 14.6 & 22.8 / 16.0 \\
JoyAI-Image-Edit & 69.1 / 23.9 & 38.1 / \underline{27.8} & 49.9 / 24.1 & 63.0 / 26.0 & 12.0 / 20.9 & 46.4 / 24.5 \\
Step1X-Edit-v1p2 & 31.3 / 19.7 & 43.6 / 24.6 & 40.1 / 19.1 & 53.3 / 18.5 & 21.1 / 21.7 & 37.9 / 20.7 \\
GPT-Image-1 & 50.0 / 14.4 & 52.8 / 12.9 & 43.7 / 13.1 & 50.6 / 13.2 & 16.3 / 13.1 & 42.7 / 13.4 \\
\midrule
\multicolumn{7}{l}{\textit{Text-instructed baselines}} \\
Qwen-Image-Edit-2511 & 63.7 / 18.9 & 71.8 / 19.7 & 54.7 / 19.9 & 67.8 / 20.0 & 54.4 / 18.9 & 62.5 / 19.4 \\
LongCat-Image-Edit & 54.6 / 17.7 & 66.9 / 19.5 & 53.4 / 20.2 & 62.7 / 19.7 & 34.7 / 16.1 & 54.5 / 18.6 \\
FLUX.2-klein-9B & 57.3 / 25.0 & 70.8 / 23.5 & 57.4 / 24.1 & 59.8 / 24.4 & 32.4 / 26.0 & 55.5 / 24.6 \\
FireRed-Image-Edit-1.0 & \underline{69.9} / 23.4 & 69.8 / 22.8 & 65.4 / 26.4 & \underline{78.9} / 25.8 & 53.0 / 21.8 & 67.4 / 24.0 \\
JoyAI-Image-Edit & 66.0 / 21.9 & 72.3 / 23.3 & 61.5 / 26.2 & 72.9 / 24.7 & 54.6 / 20.8 & 65.4 / 23.3 \\
Step1X-Edit-v1p2 & 43.3 / 18.8 & 60.8 / 19.5 & 50.6 / 19.8 & 62.5 / 19.6 & 31.9 / 17.8 & 49.8 / 19.1 \\
\midrule
\multicolumn{7}{l}{\textit{Ours}} \\
VibeEdit-base & 67.6 / \underline{32.7} & \underline{73.3} / 27.5 & \underline{66.0} / 22.1 & 72.9 / \underline{31.5} & \underline{59.1} / 20.7 & \underline{67.8} / \underline{27.0} \\
\textbf{VibeEdit} & \textbf{80.3} / \textbf{34.7} & \textbf{85.4} / \textbf{35.8} & \textbf{77.4} / \textbf{31.2} & \textbf{79.9} / \textbf{34.9} & \textbf{76.4} / \underline{27.1} & \textbf{79.9} / \textbf{32.8} \\
\bottomrule
\end{tabular}%
\vspace{-3mm}
\end{table}

\begin{table*}[t]
\centering
\begin{minipage}[t]{0.48\textwidth}
\vspace{0pt}
\caption{Effect of VAE inputs on editing performance,
with the VLM input fixed to $A$.}
\label{tab:decoupling-ablation}
\end{minipage}
\hfill
\begin{minipage}[t]{0.48\textwidth}
\vspace{0pt}
\caption{Effect of region weighting and mask dilation
on the base model.}
\label{tab:mask-loss-ablation}
\end{minipage}

\par\vspace{2pt}

\begin{minipage}[t]{0.48\textwidth}
\vspace{0pt}
\centering
\small
\renewcommand{\arraystretch}{1.12}
\begin{tabular*}{\linewidth}{@{\extracolsep{\fill}}lcc@{}}
\toprule
\textbf{VAE input} &
\textbf{VLM $\uparrow$} &
\textbf{PSNR $\uparrow$} \\
\midrule
$I$       & 24.7 & 21.3 \\
$A$       & \underline{58.6} & \underline{24.2} \\
$I,\bar C$ & \textbf{67.8} & \textbf{27.0} \\
\bottomrule
\end{tabular*}
\end{minipage}
\hfill
\begin{minipage}[t]{0.48\textwidth}
\vspace{0pt}
\centering
\small
\renewcommand{\arraystretch}{1.12}
\begin{tabular*}{\linewidth}{@{\extracolsep{\fill}}ccc@{}}
\toprule
\textbf{Weighting} &
\textbf{Dilation} &
\textbf{VLM $\uparrow$} \\
\midrule
--           & --           & \underline{66.2} \\
$\checkmark$ & --           & 64.2 \\
$\checkmark$ & $\checkmark$ & \textbf{67.8} \\
\bottomrule
\end{tabular*}
\end{minipage}

\vspace{-4mm}
\end{table*}

\subsection{Main Results}
\label{sec:main-comparison}

We compare VibeEdit with Qwen-Image-Edit-2511
\citep{wu2025qwenimage}, LongCat-Image-Edit \citep{longcat2025image},
FLUX.2-klein-9B \citep{blackforestlabs2026flux2klein},
FireRed-Image-Edit-1.0 \citep{qiao2026firered},
JoyAI-Image-Edit \citep{song2026joyai},
Step1X-Edit-v1p2 \citep{liu2025step1x}, and proprietary GPT-Image-1 \citep{openai2025gptimage}.
Baselines evaluated without an external text prompt receive the
canvas-annotated image $A$, including any rasterized notes.
We also evaluate public baselines in a text-instructed setting,
providing the source image $I$ and the full human-written benchmark
instruction as an external text prompt (21.3 words on average).
VibeEdit receives source image and canvas instruction $(I,C)$
without a separate text prompt.

\noindent\textbf{Main results.}
Table~\ref{tab:main-results} shows that all six public baselines
achieve higher overall rubric scores with external text instructions
than with the canvas-annotated image alone. Without an external text prompt, our base model achieves a VLM
rubric score comparable to that of FireRed, the highest-scoring
text-instructed baseline (67.8 versus 67.4), while achieving
higher outside-region PSNR.
The final VibeEdit configuration achieves the highest rubric score
on all five edit types and the highest overall PSNR, reaching
79.9 and 32.8 dB, respectively.

\begin{figure}[t]
    \centering
    \includegraphics[width=1.0\textwidth]{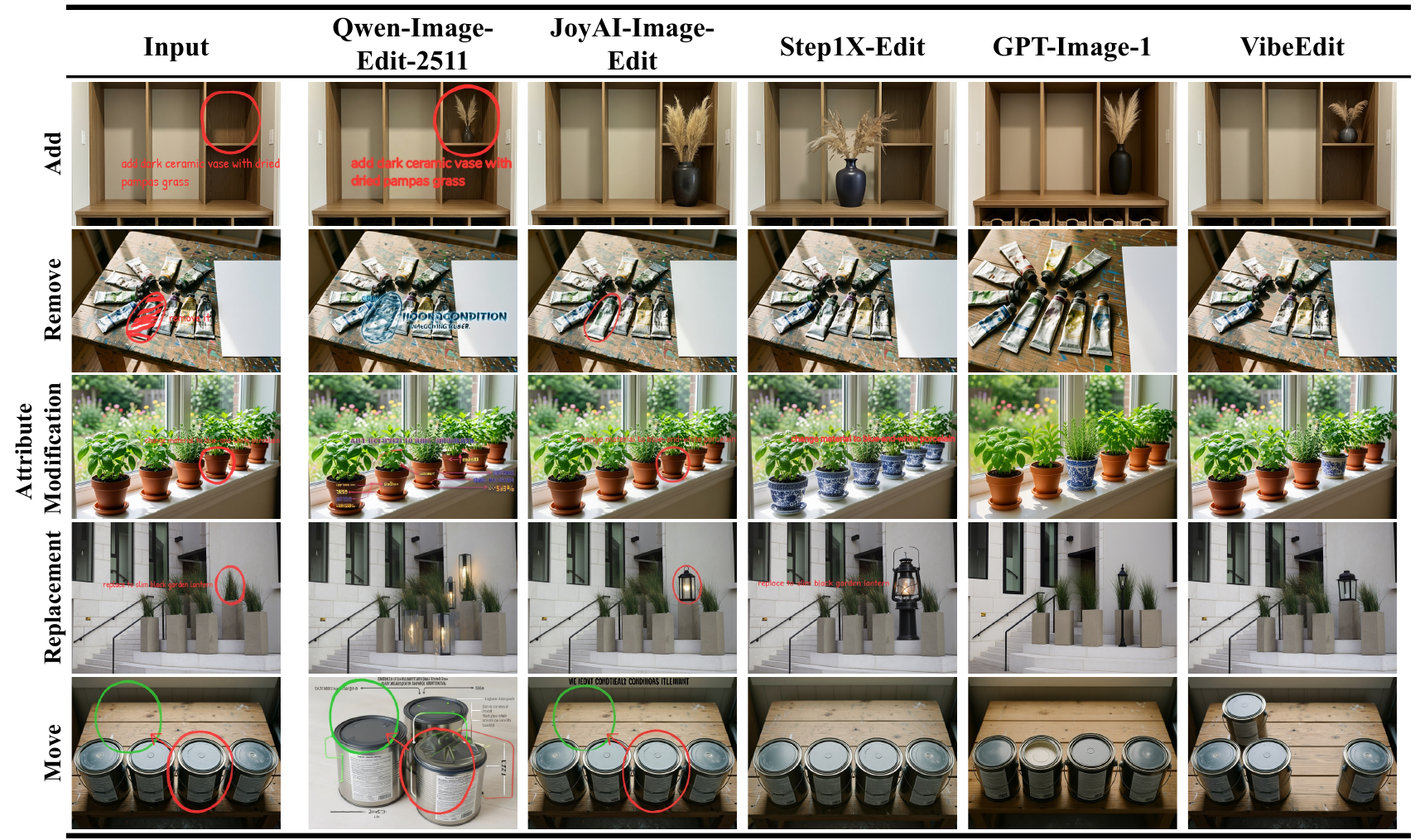}
    \vspace{-5mm}
    \caption{Qualitative comparison across all five edit types without
    separate text prompts. Baseline methods exhibit typical errors such as instruction-mark retention, target-instance confusion, unintended changes to unrelated regions, and inaccurate object relocation. Additional qualitative results are provided in Figure~\ref{fig:more-demo} in the Appendix.}
    \vspace{-4mm}
    \label{fig:qualitative-comparison}
\end{figure}

\noindent\textbf{Qualitative comparison.}
Figure~\ref{fig:qualitative-comparison} compares editing results
across all five edit types. The baselines exhibit varying errors
in instruction following and source preservation, while VibeEdit
follows the canvas instructions more consistently.

\subsection{Ablation Studies}
\label{sec:ablation}

\noindent\textbf{Layer-decoupled conditioning.}
Table~\ref{tab:decoupling-ablation} compares VAE inputs while keeping the annotated image $A$ fixed at the vision--language encoder. Encoding only the source image $I$ leaves the canvas instruction represented solely by semantic tokens and yields a rubric score of 24.7. Encoding $A$ raises the score to 58.6. Separately encoding $I$ and $\bar C$ further improves the rubric score to 67.8 and PSNR from 24.2 to 27.0 dB. These results support separately encoding the annotation layout
and the clean image.

\noindent\textbf{Region-weighted supervision.}
\label{sec:mask-loss-ablation}
As shown in Table~\ref{tab:mask-loss-ablation}, region weighting
without dilation lowers the rubric score from 66.2 to 64.2,
while combining weighting with dilation raises it to 67.8.
Dilation extends the upweighted region around object boundaries
and annotations to accommodate approximate marks and nearby
edit effects.

\noindent\textbf{RL condition selection.}
\label{sec:rl-data-selection}
Table~\ref{tab:rl-data-selection} compares selection strategies
based on rollout-reward variation. High-variance selection
achieves a slightly higher rubric score than random sampling,
at 79.9 versus 79.3, while low-variance selection performs
worse at 76.6.

\noindent\textbf{Reward composition.}
Table~\ref{tab:reward-ablation} shows that local edit quality
and outside preservation play complementary roles. Starting
from edit-success rewards alone, adding local-quality criteria
improves the rubric score but reduces outside-region PSNR,
whereas adding preservation criteria improves both metrics.
Combining all three rubric groups achieves the best results,
with a rubric score of 79.9 and PSNR of 32.8 dB.

\begin{table*}[t]
\centering
\begin{minipage}[t]{0.42\textwidth}
\vspace{0pt}
\caption{Effect of RL condition selection based on
variation in rollout rewards.}
\label{tab:rl-data-selection}
\end{minipage}
\hfill
\begin{minipage}[t]{0.55\textwidth}
\vspace{0pt}
\caption{Effect of RL reward components on edit quality
and source preservation.}
\label{tab:reward-ablation}
\end{minipage}

\par\vspace{2pt}

\begin{minipage}[t]{0.42\textwidth}
\vspace{0pt}
\centering
\small
\renewcommand{\arraystretch}{1.12}
\setlength{\tabcolsep}{3pt}
\begin{tabular*}{\linewidth}{@{\extracolsep{\fill}}lc@{}}
\toprule
\textbf{Selection Strategy} & \textbf{VLM $\uparrow$} \\
\midrule
High-variance & \textbf{79.9} \\[0.70\baselineskip]
Random       & \underline{79.3} \\[0.70\baselineskip]
Low-variance & 76.6 \\[0.70\baselineskip]
\bottomrule
\end{tabular*}
\end{minipage}
\hfill
\begin{minipage}[t]{0.55\textwidth}
\vspace{0pt}
\raggedleft
\small
\renewcommand{\arraystretch}{1.12}
\setlength{\tabcolsep}{5pt}
\begin{tabular}{@{}ccccc@{}}
\toprule
\begin{tabular}[c]{@{}c@{}}
    \textbf{Edit}\\
    \textbf{Success}
\end{tabular}
&
\begin{tabular}[c]{@{}c@{}}
    \textbf{Outside}\\
    \textbf{Preservation}
\end{tabular}
&
\begin{tabular}[c]{@{}c@{}}
    \textbf{Local Edit}\\
    \textbf{Quality}
\end{tabular}
&
\textbf{VLM $\uparrow$}
&
\textbf{PSNR $\uparrow$}
\\
\midrule
$\checkmark$ & --           & --           & 69.2 & 28.6 \\
$\checkmark$ & $\checkmark$ & --           & \underline{74.9} & \underline{29.4} \\
$\checkmark$ & --           & $\checkmark$ & 72.4 & 28.0 \\
$\checkmark$ & $\checkmark$ & $\checkmark$ & \textbf{79.9} & \textbf{32.8} \\
\bottomrule
\end{tabular}
\end{minipage}
\vspace{-8mm}
\end{table*}

\section{Conclusion}
\label{sec:conclusion}

We introduced a new image editing interface in which spatial marks
and optional short notes form canvas instructions, specifying
where to edit and what to change. We developed VibeEdit to support
five editing operations through this interface without a separate
text prompt. Using 1.55 million source--target edit pairs, we trained
VibeEdit with layer-decoupled conditioning, region-weighted supervised learning,
and rubric-guided RL. On an independent, human-curated 419-case
benchmark emphasizing target selection among similar objects,
VibeEdit outperforms the evaluated baselines in overall VLM
rubric score and outside-region PSNR.

\subsection*{AI use statement}
The authors developed the research ideas and methodology without
generative AI assistance. We used GPT models for object selection
and edit-description generation, and generative image models to
synthesize training pairs, as described in Section~\ref{sec:dataset}.
GPT models also provided rubric-based rewards during reinforcement
learning and evaluated editing outputs. We also used language models
to improve the clarity and readability of the manuscript.
The authors take responsibility for all content, claims, and
artifacts in this work, including those produced with AI assistance.

\subsection*{Ethics statement}
Our data include images from publicly available datasets and
AI-generated images.

\subsection*{Reproducibility statement}
To ensure reproducibility, Sections~\ref{sec:dataset},
\ref{sec:method}, and~\ref{sec:experimental-setup} describe
data construction, the model and training objectives, and
the training, inference, and evaluation settings.
The appendix provides additional data details, the RL policy
update, data-selection rules, and evaluation prompts and rubrics.

\newpage

\bibliography{iclr2027_conference}

@inproceedings{brooks2023instructpix2pix,
  title={Instructpix2pix: Learning to follow image editing instructions},
  author={Brooks, Tim and Holynski, Aleksander and Efros, Alexei A},
  booktitle={2023 IEEE/CVF Conference on Computer Vision and Pattern Recognition (CVPR)},
  pages={18392--18402},
  year={2023},
  organization={IEEE}
}

@article{zhang2023magicbrush,
  title={Magicbrush: A manually annotated dataset for instruction-guided image editing},
  author={Zhang, Kai and Mo, Lingbo and Chen, Wenhu and Sun, Huan and Su, Yu},
  journal={Advances in neural information processing systems},
  volume={36},
  pages={31428--31449},
  year={2023}
}

@article{zhao2024ultraedit,
  title={Ultraedit: Instruction-based fine-grained image editing at scale},
  author={Zhao, Haozhe and Ma, Xiaojian and Chen, Liang and Si, Shuzheng and Wu, Rujie and An, Kaikai and Yu, Peiyu and Zhang, Minjia and Li, Qing and Chang, Baobao},
  journal={Advances in Neural Information Processing Systems},
  volume={37},
  pages={3058--3093},
  year={2024}
}

@article{wu2025qwenimage,
  title={Qwen-image technical report},
  author={Wu, Chenfei and Li, Jiahao and Zhou, Jingren and Lin, Junyang and Gao, Kaiyuan and Yan, Kun and Yin, Sheng-ming and Bai, Shuai and Xu, Xiao and Chen, Yilei and others},
  journal={arXiv preprint arXiv:2508.02324},
  year={2025}
}

@article{zhao2026qwenimage20,
  title={Qwen-image-2.0 technical report},
  author={Zhao, Bing and Wu, Chenfei and Li, Deqing and Meng, Hao and Li, Jiahao and Zhang, Jie and Zhou, Jingren and Lin, Junyang and Gao, Kaiyuan and Cao, Kuan and others},
  journal={arXiv preprint arXiv:2605.10730},
  year={2026}
}

@article{cao2025hunyuanimage,
  title={HunyuanImage 3.0 Technical Report},
  author={Cao, Siyu and Chen, Hangting and Chen, Peng and Cheng, Yiji and Cui, Yutao and Deng, Xinchi and Dong, Ying and Gong, Kipper and Gu, Tianpeng and Gu, Xiusen and others},
  journal={arXiv preprint arXiv:2509.23951},
  year={2025}
}

@article{zimage2025,
  title={Z-image: An efficient image generation foundation model with single-stream diffusion transformer},
  author={Cai, Huanqia and Cao, Sihan and Du, Ruoyi and Gao, Peng and Hao, Aiming and Hoi, Steven and Hou, Zhaohui and Huang, Shijie and Jiang, Dengyang and Jiang, Yuming and others},
  journal={arXiv preprint arXiv:2511.22699},
  year={2025}
}

@article{seedream2025seedream40,
  title={Seedream 4.0: Toward next-generation multimodal image generation},
  author={Seedream, Team and Chen, Yunpeng and Gao, Yu and Gong, Lixue and Guo, Meng and Guo, Qiushan and Guo, Zhiyao and Hou, Xiaoxia and Huang, Weilin and Huang, Yixuan and others},
  journal={arXiv preprint arXiv:2509.20427},
  year={2025}
}

@article{chen2026lens,
  title={Lens: Rethinking Training Efficiency for Foundational Text-to-Image Models},
  author={Chen, Dong and Wei, Fangyun and Wan, Ziyu and Chen, Dongdong and Zhang, Jiawei and Zhao, Jinjing and Zhang, Sirui and Yue, Yang and Liang, Zhiyang and Guo, Baining and others},
  journal={arXiv preprint arXiv:2605.21573},
  year={2026}
}

@inproceedings{xiao2024florence2,
  title={Florence-2: Advancing a unified representation for a variety of vision tasks},
  author={Xiao, Bin and Wu, Haiping and Xu, Weijian and Dai, Xiyang and Hu, Houdong and Lu, Yumao and Zeng, Michael and Liu, Ce and Yuan, Lu},
  booktitle={2024 IEEE/CVF Conference on Computer Vision and Pattern Recognition (CVPR)},
  pages={4818--4829},
  year={2024},
  organization={IEEE}
}

@inproceedings{carion2025sam3,
  title={Sam 3: Segment anything with concepts},
  author={Carion, Nicolas and Gustafson, Laura and Hu, Yuan-Ting and Debnath, Shoubhik and Hu, Ronghang and Suris Coll-Vinent, Didac and Ryali, Chaitanya and Alwala, Kalyan Vasudev and Khedr, Haitham and Huang, Andrew and others},
  booktitle={International conference on learning representations},
  volume={2026},
  pages={138846--138923},
  year={2026}
}

@inproceedings{zhao2026objectclear,
  title={Precise object and effect removal with adaptive target-aware attention},
  author={Zhao, Jixin and Wang, Zhouxia and Yang, Peiqing and Zhou, Shangchen},
  booktitle={Proceedings of the IEEE/CVF Conference on Computer Vision and Pattern Recognition},
  pages={19370--19379},
  year={2026}
}

@misc{blackforestlabs2024fluxfill,
  title={Introducing {FLUX.1} Tools},
  author={{Black Forest Labs}},
  year={2024},
  howpublished={\url{https://bfl.ai/blog/24-11-21-tools}}
}

@inproceedings{pan2023draggan,
  title={Drag your gan: Interactive point-based manipulation on the generative image manifold},
  author={Pan, Xingang and Tewari, Ayush and Leimk{\"u}hler, Thomas and Liu, Lingjie and Meka, Abhimitra and Theobalt, Christian},
  booktitle={ACM SIGGRAPH 2023 conference proceedings},
  pages={1--11},
  year={2023}
}

@inproceedings{shi2023dragdiffusion,
  title={Dragdiffusion: Harnessing diffusion models for interactive point-based image editing},
  author={Shi, Yujun and Xue, Chuhui and Liew, Jun Hao and Pan, Jiachun and Yan, Hanshu and Zhang, Wenqing and Tan, Vincent YF and Bai, Song},
  booktitle={2024 IEEE/CVF Conference on Computer Vision and Pattern Recognition (CVPR)},
  pages={8839--8849},
  year={2024},
  organization={IEEE}
}

@inproceedings{liu2025magicquill,
  title={Magicquill: An intelligent interactive image editing system},
  author={Liu, Zichen and Yu, Yue and Ouyang, Hao and Wang, Qiuyu and Cheng, Ka Leong and Wang, Wen and Liu, Zhiheng and Chen, Qifeng and Shen, Yujun},
  booktitle={2025 IEEE/CVF Conference on Computer Vision and Pattern Recognition (CVPR)},
  pages={13072--13082},
  year={2025},
  organization={IEEE}
}

@inproceedings{liu2026magicquillv2,
  title={MagicQuill V2: Precise and Interactive Image Editing with Layered Visual Cues},
  author={Liu, Zichen and Yu, Yue and Ouyang, Hao and Wang, Qiuyu and Ma, Shuailei and Cheng, Ka Leong and Wang, Wen and Bai, Qingyan and Zhang, Yuxuan and Zeng, Yanhong and others},
  booktitle={Proceedings of the IEEE/CVF Conference on Computer Vision and Pattern Recognition},
  pages={22467--22477},
  year={2026}
}

@inproceedings{liu2026interedit,
  title={Inter-Edit: First Benchmark for Interactive Instruction-Based Image Editing},
  author={Liu, Delong and Hou, Haotian and Hou, Zhaohui and Huang, Zhiyuan and Han, Shihao and Zhan, Mingjie and Zhao, Zhicheng and Su, Fei},
  booktitle={Proceedings of the IEEE/CVF Conference on Computer Vision and Pattern Recognition},
  pages={37290--37300},
  year={2026}
}

@article{xu2026fineedit,
  title={FineEdit: Fine-Grained Image Edit with Bounding Box Guidance},
  author={Xu, Haohang and Liu, Lin and Zhang, Zhibo and Cong, Rong and Zhang, Xiaopeng and Tian, Qi},
  journal={arXiv preprint arXiv:2604.10954},
  year={2026}
}

@article{xu2026scribble,
  title={Rethinking Scribble-Guided Image Editing: Generalization, Instruction Adherence, and Multi-Tasking},
  author={Xu, Mingyi and Lin, Jinpeng and Zhou, Min and Ge, Tiezheng and Zeng, Ming},
  journal={arXiv preprint arXiv:2605.25568},
  year={2026}
}

@article{zhang2026vibe,
  title={How Well Do Models Follow Visual Instructions? VIBE: A Systematic Benchmark for Visual Instruction-Driven Image Editing},
  author={Zhang, Huanyu and Bai, Xuehai and Li, Chengzu and Liang, Chen and Tian, Haochen and Li, Haodong and An, Ruichuan and Zhang, Yifan and Korhonen, Anna and Zhang, Zhang and others},
  journal={arXiv preprint arXiv:2602.01851},
  year={2026}
}

@inproceedings{black2023ddpo,
  title={Training diffusion models with reinforcement learning},
  author={Black, Kevin and Janner, Michael and Du, Yilun and Kostrikov, Ilya and Levine, Sergey},
  booktitle={International Conference on Learning Representations},
  volume={2024},
  pages={4965--4987},
  year={2024}
}

@inproceedings{liu2025flowgrpo,
  title     = {Flow-GRPO: Training Flow Matching Models via Online RL},
  author    = {Jie Liu and Gongye Liu and Jiajun Liang and Yangguang Li
               and Jiaheng Liu and Xintao Wang and Pengfei Wan
               and Di Zhang and Wanli Ouyang},
  booktitle = {Advances in Neural Information Processing Systems},
  volume    = {38},
  year      = {2025},
  doi       = {10.52202/085713-1362}
}

@article{xue2025dancegrpo,
  title={Dancegrpo: Unleashing grpo on visual generation},
  author={Xue, Zeyue and Wu, Jie and Gao, Yu and Kong, Fangyuan and Zhu, Lingting and Chen, Mengzhao and Liu, Zhiheng and Liu, Wei and Guo, Qiushan and Huang, Weilin and others},
  journal={arXiv preprint arXiv:2505.07818},
  year={2025}
}

@article{zhang2026opgrpo,
  title={OP-GRPO: Efficient Off-Policy GRPO for Flow-Matching Models},
  author={Zhang, Liyu and Li, Kehan and Han, Tingrui and Zhao, Tao and Sheng, Yuxuan and He, Shibo and Li, Chao},
  journal={arXiv preprint arXiv:2604.04142},
  year={2026}
}

@inproceedings{zheng2025diffusionnft,
  title={Diffusionnft: Online diffusion reinforcement with forward process},
  author={Zheng, Kaiwen and Chen, Huayu and Ye, Haotian and Wang, Haoxiang and Zhang, Qinsheng and Jiang, Kai and Su, Hang and Ermon, Stefano and Zhu, Jun and Liu, Ming-Yu},
  booktitle={International Conference on Learning Representations},
  volume={2026},
  pages={134129--134150},
  year={2026}
}

@article{feng2025rubricrl,
  title={Rubricrl: Simple generalizable rewards for text-to-image generation},
  author={Feng, Xuelu and Li, Yunsheng and Wan, Ziyu and Gao, Zixuan and Yuan, Junsong and Chen, Dongdong and Qiao, Chunming},
  journal={arXiv preprint arXiv:2511.20651},
  year={2025}
}

@article{longcat2025image,
  title={Longcat-image technical report},
  author={Team, Meituan LongCat and Ma, Hanghang and Tan, Haoxian and Huang, Jiale and Wu, Junqiang and He, Jun-Yan and Gao, Lishuai and Xiao, Songlin and Wei, Xiaoming and Ma, Xiaoqi and others},
  journal={arXiv preprint arXiv:2512.07584},
  year={2025}
}

@misc{blackforestlabs2026flux2klein,
  title={{FLUX.2 [klein]}},
  author={{Black Forest Labs}},
  year={2026},
  howpublished={\url{https://bfl.ai/models/flux-2-klein}}
}

@article{qiao2026firered,
  title={Firered-image-edit-1.0 technical report},
  author={Team, Super Intelligence and Qiao, Changhao and Hui, Chao and Li, Chen and Wang, Cunzheng and Song, Dejia and Zhang, Jiale and Li, Jing and Xiang, Qiang and Wang, Runqi and others},
  journal={arXiv preprint arXiv:2602.13344},
  year={2026}
}

@article{song2026joyai,
  title={JoyAI-Image: Awaking Spatial Intelligence in Unified Multimodal Understanding and Generation},
  author={Song, Lin and Li, Wenbo and Ma, Guoqing and Tang, Wei and Wang, Bo and Zhang, Yuan and Yang, Yijun and Xiao, Yicheng and Liu, Jianhui and Zhang, Yanbing and others},
  journal={arXiv preprint arXiv:2605.04128},
  year={2026}
}

@article{liu2025step1x,
  title={Step1x-edit: A practical framework for general image editing},
  author={Liu, Shiyu and Han, Yucheng and Xing, Peng and Yin, Fukun and Wang, Rui and Cheng, Wei and Liao, Jiaqi and Wang, Yingming and Fu, Honghao and Han, Chunrui and others},
  journal={arXiv preprint arXiv:2504.17761},
  year={2025}
}

@misc{openai2025gptimage,
  author       = {{OpenAI}},
  title        = {{GPT-Image-1}},
  year         = {2025},
  howpublished = {\url{https://developers.openai.com/api/docs/models/gpt-image-1}},
  note         = {OpenAI API model documentation}
}

@misc{google2025nanobanana,
  author = {{Google}},
  title  = {{Gemini 2.5 Flash Image (Nano Banana)}},
  year   = {2025},
  url    = {https://ai.google.dev/gemini-api/docs/models/gemini-2.5-flash-image}
}

@article{hu2021lora,
  title={Lora: Low-rank adaptation of large language models},
  author={Hu, Edward J and Shen, Yelong and Wallis, Phillip and Allen-Zhu, Zeyuan and Li, Yuanzhi and Wang, Shean and Wang, Lu and Chen, Weizhu},
  journal={arXiv preprint arXiv:2106.09685},
  year={2021}
}

@misc{openai2026gpt56,
  author       = {{OpenAI}},
  title        = {GPT-5.6: Frontier Intelligence That Scales with Your Ambition},
  year         = {2026},
  howpublished = {\url{https://openai.com/index/gpt-5-6/}},
}

@misc{openai2026gpt54,
  author       = {{OpenAI}},
  title        = {Introducing {GPT-5.4}},
  year         = {2026},
  howpublished = {\url{https://openai.com/index/introducing-gpt-5-4/}},
}

@misc{bai2025qwen25vltechnicalreport,
      title={Qwen2.5-VL Technical Report}, 
      author={Shuai Bai and Keqin Chen and Xuejing Liu and Jialin Wang and Wenbin Ge and Sibo Song and Kai Dang and Peng Wang and Shijie Wang and Jun Tang and Humen Zhong and Yuanzhi Zhu and Mingkun Yang and Zhaohai Li and Jianqiang Wan and Pengfei Wang and Wei Ding and Zheren Fu and Yiheng Xu and Jiabo Ye and Xi Zhang and Tianbao Xie and Zesen Cheng and Hang Zhang and Zhibo Yang and Haiyang Xu and Junyang Lin},
      year={2025},
      eprint={2502.13923},
      archivePrefix={arXiv},
      primaryClass={cs.CV},
      url={https://arxiv.org/abs/2502.13923}, 
}
\bibliographystyle{iclr2027_conference}

\clearpage
\appendix
\section{Supplementary Experiments}
\label{sec:supplementary-experiments}

We further examine task composition, annotation color, font diversity,
removal instructions, and LoRA configuration. We also validate the
automatic evaluation judge against human annotations. VLM rubric
scores and outside-region PSNR follow the definitions in
Section~\ref{sec:experimental-setup}; evaluation subsets and
human-evaluation settings are specified below.

\noindent\textbf{Joint and single-task training.}
Table~\ref{tab:supp-joint-training} compares single-task base models
with a model trained jointly on all five editing tasks.
Joint training improves the VLM rubric score over the corresponding
single-task models on addition, attribute modification, and
movement, suggesting positive transfer across editing operations.
Although removal-only and replacement-only models perform better
on their respective tasks, the results support learning multiple
editing operations within a shared model.

\begin{table}[!h]
\vspace{-2mm}
\caption{Effect of task composition in supervised learning. All models are evaluated
on the full benchmark. Entries report VLM rubric scores (\%);
higher is better. Bold and underlined values indicate the best
and second-best results in each column.}
\vspace{2mm}
\label{tab:supp-joint-training}
\centering
\begin{tabular}{lcccccc}
\toprule
\textbf{Training Data} & \textbf{Add} & \textbf{Remove} &
\textbf{Attribute} & \textbf{Replace} &
\textbf{Move} & \textbf{Overall} \\
\midrule
Add only
& 42.7 & 45.0 & \underline{62.7} & 46.6 & 28.8 & 45.2 \\
Remove only
& 28.7 & \textbf{81.4} & 40.7 & 50.7 & 37.6 & 47.8 \\
Attribute only
& 60.1 & 35.0 & 59.8 & 55.4 & 19.1 & 45.9 \\
Replace only
& \underline{61.2} & 53.4 & 60.1 & \textbf{77.6} & 27.9 & \underline{56.1} \\
Move only
& 23.8 & 31.3 & 44.7 & 34.3 & \underline{55.9} & 38.0 \\
\textbf{All five tasks}
& \textbf{67.6} & \underline{73.3} & \textbf{66.0}
& \underline{72.9} & \textbf{59.1} & \textbf{67.8} \\
\bottomrule
\end{tabular}
\end{table}

\noindent\textbf{Annotation-color generalization.}
Table~\ref{tab:color-generalization} evaluates changes to the primary
stroke color, which is red during training. At evaluation, we use
blue, yellow, or a color sampled independently for each case from
red, blue, and yellow. Movement destination circles remain
green in all settings. Relative to red strokes, the rubric score
changes by at most 0.6 points and PSNR by at most 0.2 dB,
indicating stable performance under these color variations.

\noindent\textbf{Font diversity.}
Table~\ref{tab:supp-font-user-study} compares models trained with
a single font or multiple fonts during supervised learning. Participants create
canvas instructions in their own handwriting and compare the
outputs from the two models. Across 200 pairwise human preference
comparisons, the model trained with multiple fonts achieves a
higher Elo score.

\begin{table}[!h]
\centering
\begin{minipage}[b]{0.54\textwidth}
\centering
\vspace{-2mm}
\caption{Robustness to changes in primary stroke color.
Training uses red strokes; movement destination circles remain
green at evaluation. ``Random'' samples red, blue, or yellow for each case.}
\vspace{2mm}
\label{tab:color-generalization}
\small
\renewcommand{\arraystretch}{1.12}
\begin{tabular}{lcc}
\toprule
\textbf{Primary Stroke Color} & \textbf{VLM $\uparrow$} &
\textbf{PSNR $\uparrow$} \\
\midrule
Red & \textbf{79.9} & \textbf{32.8} \\
Blue & 79.4 & \underline{32.6} \\
Yellow & 79.3 & \underline{32.6} \\
Random & \underline{79.5} & \underline{32.6} \\
\bottomrule
\end{tabular}
\end{minipage}
\hfill
\begin{minipage}[b]{0.43\textwidth}
\centering
\vspace{-2mm}
\caption{Effect of font diversity during supervised learning on human preferences
for edits guided by user-written canvas instructions.
Elo scores summarize 200 pairwise comparisons.}
\vspace{2mm}
\label{tab:supp-font-user-study}
\small
\renewcommand{\arraystretch}{1.12}
\begin{tabular}{lc}
\toprule
\textbf{Training Fonts} & \textbf{Human Elo $\uparrow$} \\
\midrule
Single font & 1382.9 \\
Multiple fonts & \textbf{1617.1} \\
\bottomrule
\end{tabular}
\end{minipage}
\end{table}

\begin{table}[h]
\vspace{-4mm}
\caption{Comparison of removal instruction designs. Four models are independently trained on the same removal data under supervised learning and evaluated on the removal subset.
 All designs circle the target;
``w/ text'' adds  ``remove it'' to the X mark or scribble.}
\vspace{2mm}
\label{tab:supp-remove-prompts}
\centering
\renewcommand{\arraystretch}{1.12}
\begin{tabular}{lcc}
\toprule
\textbf{Canvas Instruction} & \textbf{VLM $\uparrow$} &
\textbf{PSNR $\uparrow$} \\
\midrule
X mark & 58.4 & 27.0 \\
X mark w/ text & \underline{78.9} & \underline{30.8} \\
Scribble & 55.4 & 27.6 \\
Scribble w/ text & \textbf{81.4} & \textbf{36.1} \\
\bottomrule
\end{tabular}
\end{table}

\begin{figure}[t]
\centering
\includegraphics[width=0.9\textwidth]{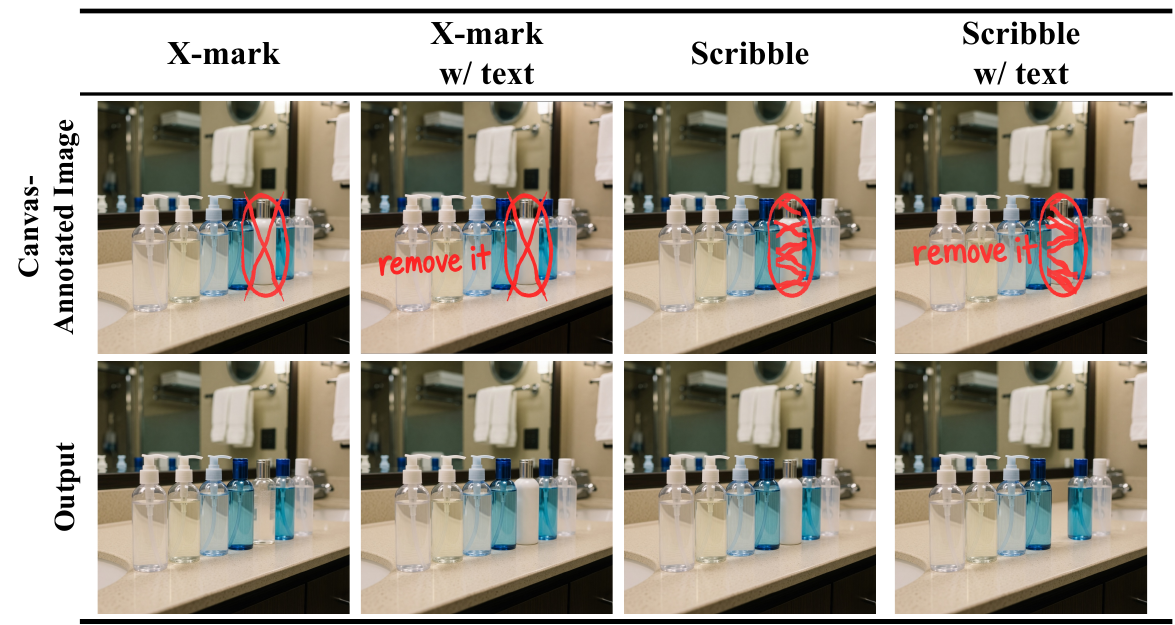}
\vspace{-2mm}
\caption{Four removal instruction designs and outputs from their
corresponding base models. Each design uses a circled X mark or
scribble, with or without a short handwritten text cue.}
\vspace{-4mm}
\label{fig:supp-remove-prompt-comparison}
\end{figure}

\noindent\textbf{Canvas instructions for removal.}
Figure~\ref{fig:supp-remove-prompt-comparison} illustrates four
instruction designs that vary the deletion mark and the presence
of a short text cue. Each design circles the target object and
uses either an X mark or a scribble. We train four separate models
by supervised learning on the same removal data, using a different annotation
design for each model, and evaluate them on the removal subset.

Table~\ref{tab:supp-remove-prompts} shows that models trained
with a text cue achieve higher rubric scores and outside-region
PSNR for both deletion marks. The circled scribble with text
performs best, suggesting that an explicit deletion cue helps
the model interpret the requested operation. We adopt the
circled scribble in our default removal instructions and
include ``remove it'' during supervised learning.

\noindent\textbf{Human validation of the VLM judge.}
Table~\ref{tab:human-evaluation} compares GPT-5.6-sol with
majority-vote human labels on 100 randomly sampled VibeEdit outputs.
Human raters and the automatic judge answer the same 947 binary
rubric questions. The pooled pass rates are close: 76.6\% for
humans and 77.0\% for GPT-5.6-sol. Question-level agreement reaches
93.0\% accuracy, 95.5\% F1, and Cohen's $\kappa=0.805$,
indicating close agreement with human judgments on these outputs.
These pass rates are pooled across questions.

\noindent\textbf{LoRA module ablation.}
Table~\ref{tab:supp-lora} compares the DiT modules adapted with
LoRA while keeping the rank, learning rate, training data,
and optimization budget fixed. Adapting both attention and
modulation improves the rubric score and PSNR over attention-only
adaptation. Further adapting the MLP lowers both scores despite
increasing the number of trainable parameters.

\begin{table}[h]
\vspace{-4mm}
\caption{Validation of GPT-5.6-sol against majority-vote human
labels on 100 randomly sampled VibeEdit outputs. Pass rates
are pooled over 947 matched binary questions. Accuracy, F1,
and Cohen's $\kappa$ measure question-level agreement.}
\vspace{2mm}
\label{tab:human-evaluation}
\centering
\renewcommand{\arraystretch}{1.12}
\begin{tabular}{lcc}
\toprule
\multicolumn{3}{c}{\textit{Pooled rubric pass rates}} \\
\midrule
\textbf{Edit Model} & \textbf{Human (\%)} &
\textbf{GPT-5.6-sol (\%)} \\
\midrule
VibeEdit & 76.6 & 77.0 \\
\midrule
\multicolumn{3}{c}{\textit{Agreement with majority-vote human labels}} \\
\midrule
\textbf{Judge} & \textbf{Accuracy / F1 (\%) $\uparrow$} &
\textbf{Cohen's $\kappa$ $\uparrow$} \\
\midrule
GPT-5.6-sol & 93.0 / 95.5 & 0.805 \\
\bottomrule
\end{tabular}
\end{table}

\begin{table}[!ht]
\vspace{-4mm}
\caption{Effect of the DiT modules adapted with LoRA.
All configurations use the same rank, learning rate,
training data, and optimization budget.}
\vspace{2mm}
\label{tab:supp-lora}
\centering
\renewcommand{\arraystretch}{1.12}
\begin{tabular}{cccrcc}
\toprule
\textbf{Attention} & \textbf{Modulation} & \textbf{MLP} &
\textbf{Trainable Params.} & \textbf{VLM $\uparrow$} &
\textbf{PSNR $\uparrow$} \\
\midrule
$\checkmark$ & & & 377.5M & \underline{64.1} & \underline{26.6} \\
$\checkmark$ & $\checkmark$ & & 707.8M & \textbf{67.8} & \textbf{27.0} \\
$\checkmark$ & $\checkmark$ & $\checkmark$ & 943.7M & 60.3 & 22.7 \\
\bottomrule
\end{tabular}
\end{table}

\section{DiffusionNFT Policy Update}
\label{sec:nft-update}

Our forward-process policy update follows DiffusionNFT, using
the structured reward and normalization in
Section~\ref{sec:method-rl}. For each condition, the old policy
$\pi_{\mathrm{old}}$ generates $K$ rollouts, indexed by
$k\in\{1,\ldots,K\}$. We re-noise each final rollout latent
$z_0^k$ to obtain $z_t^k$. The current and old policies then
predict velocities $v_\theta$ and $v_{\mathrm{old}}$ from the
same noisy latent, timestep, and editing condition.
The implicit positive and negative predictions are
\begin{equation}
    v_\theta^+=(1-\beta)v_{\mathrm{old}}+\beta v_\theta,
    \qquad
    v_\theta^-=(1+\beta)v_{\mathrm{old}}-\beta v_\theta,
\label{eq:nft-branches}
\end{equation}
where $\beta$ controls the deviation of each branch from
the old policy prediction.
For a velocity prediction $v$ on rollout $k$, let
$\hat z_0^k(v)=z_t^k-\sigma_t v$ and define the self-normalized reconstruction
loss
\begin{equation}
    \ell(v;z_0^k)=
    \frac{\operatorname{mean}\!\left[
    \left(\hat z_0^k(v)-z_0^k\right)^2\right]}
    {\operatorname{sg}\!\left[
    \operatorname{mean}|\hat z_0^k(v)-z_0^k|\right]},
\end{equation}
where $\operatorname{sg}$ stops gradients. The resulting objective is
\begin{equation}
    \mathcal{L}_{\mathrm{RL}}
    =\frac{a_{\max}}{\beta}\,
    \mathbb{E}_{k,t}
    \left[\rho_k\ell(v_\theta^+;z_0^k)
    +(1-\rho_k)\ell(v_\theta^-;z_0^k)\right]
    +\eta\|v_\theta-v_{\mathrm{ref}}\|_2^2,
\label{eq:nft-loss}
\end{equation}
where $v_{\mathrm{ref}}$ is the velocity prediction of the frozen
reference model and $\eta$ controls the regularization strength.
The reward-derived weight $\rho_k$, defined in
Equation~\ref{eq:rl-optimality}, balances the two reconstruction
losses: within each condition, higher-reward rollouts place more
weight on the positive branch, while lower-reward rollouts place
more weight on the negative branch. We train a new LoRA adapter
for the current policy and softly update $\pi_{\mathrm{old}}$
toward it after each training round, keeping the reference model
fixed. We use $K=12$ rollouts per condition, $a_{\max}=5$,
$\delta=10^{-4}$, $\beta=1$, and $\eta=10^{-4}$.

\section{Additional Data Details}
\label{sec:supp-data}

\subsection{Canvas Instruction Synthesis}
\label{sec:in-canvas-data}

\noindent\textbf{Stored records and direction sampling.}
Each record stores a source--target image pair, object masks,
and structured edit fields. For data sources that support
bidirectional sampling, the loader selects either direction
with equal probability. Reversing an edit swaps the source and
target images and updates the instruction fields accordingly.
For addition and removal, this exchanges the object-present
and object-absent images. For attribute modification, the
original attribute value becomes the new target; for replacement,
the original object's description specifies the replacement.
Movement reversal swaps the source and destination masks
and reverses the arrow. Both directions share one stored pair.

\noindent\textbf{Online rendering.}
After selecting the edit direction, the loader applies the
same crop and resize to both images and their masks.
The renderer uses each transformed mask to locate the object
and determine its spatial extent, then draws a circle around
the masked region with a margin beyond its boundary.
The circle indicates the edit region without tracing the
exact object shape. We add operation-specific marks and
place text notes near the circles as needed, forming the
canvas instruction $C$. The same annotations are rendered
on the source image $I$ to form $A=I\oplus C$ and on an
RGB $(128,128,128)$ background to form $\bar C$, ensuring
consistent appearance and spatial layout across the
vision--language and VAE pathways. The target image
$I_{\mathrm{edit}}$ remains free of annotations.

\noindent\textbf{Operation-specific instructions.}
Table~\ref{tab:annotation-grammar} summarizes how these circles
are combined with operation-specific marks and text notes
to express each edit type. For addition, the mask from
the object-present image identifies the insertion region in
the object-absent source. The primary stroke color is red;
movement uses a red source circle and a green destination
circle connected by a directed arrow. Removal uses a circled
scribble, accompanied by ``remove it'' in the default supervised learning configuration. 
Text notes are rendered as
part of the canvas annotations and are never supplied as a
separate text prompt.

\begin{table}[t]
\centering
\small
\vspace{-4mm}
\renewcommand{\arraystretch}{1.12}
\caption{Canvas instruction grammar for each operation.
Braced placeholders are filled from the structured edit fields.
Text templates reproduce the renderer's wording.}
\vspace{2mm}
\label{tab:annotation-grammar}
\setlength\tabcolsep{4pt}
\begin{tabular}{p{0.22\textwidth}p{0.31\textwidth}p{0.41\textwidth}}
\toprule
\textbf{Operation} & \textbf{Spatial Marks} & \textbf{Rasterized Note} \\
\midrule
Addition & Circle around the insertion region &
\texttt{add \{object\}} \\
Removal & Circle and scribble over the object &
\texttt{remove it} \\
Attribute Modification & Circle around the selected object &
\texttt{change \{attribute\} to \{value\}} \\
Replacement & Circle around the selected object &
\texttt{replace to \{description\}} \\
Movement & Source and destination circles connected by a directed arrow &
None \\
\bottomrule
\end{tabular}
\vspace{-4mm}
\end{table}

\noindent\textbf{Rendering variation.}
We randomly vary the spacing between each circle and the
object mask boundary from 0.25 to 0.45 times the mask's
equivalent radius, and sample stroke opacity in $[0.82,1.0]$.
We also vary the circle shape, stroke width, and overlap
between the stroke's start and end points. For text notes,
we sample handwriting and print typefaces, adjust font size
and line wrapping to the available space, and place the text
near the selected region while avoiding overlap where possible.
These variations produce different canvas instructions for
the same edit pair across training iterations. The annotation
alpha mask records the locations of rendered strokes and text
and is used to construct $M_{\mathrm{ann}}$ for the
region-weighted loss in Section~\ref{sec:method-sft}.

\subsection{Dataset Statistics}
\label{sec:data-statistics}

Table~\ref{tab:data-statistics} summarizes the supervised learning corpus, RL set,
and VibeEdit benchmark by edit type. Supervised learning counts refer to stored
image pairs and exclude online reversals and randomized canvas
instruction renderings. Each bidirectional addition/removal record
is counted once. The RL set contains 3,520 conditions, with 704
per edit type. The VibeEdit benchmark contains 419 cases across
the five edit types. To evaluate models that require text
instructions, each case includes a human-written external text
prompt, averaging 21.26 words across the benchmark.

\begin{table}[h]
    \centering
    \small
    \vspace{-4mm}
    \caption{Composition of the supervised learning corpus, RL set,
    and VibeEdit benchmark. Average word counts refer to the external
    text prompts used for text-instructed benchmark baselines.}
    \label{tab:data-statistics}
    \begin{tabular}{llrr}
        \toprule
        \textbf{Resource} & \textbf{Edit Category} &
        \textbf{Count} & \textbf{Avg. Words} \\
        \midrule
        \multirow{5}{*}{Supervised Learning Corpus}
            & Addition/Removal & 435,496 & -- \\
            & Attribute Modification & 458,154 & -- \\
            & Object Replacement & 393,440 & -- \\
            & Object Movement & 266,972 & -- \\
            & \textit{Total} & \textit{1,554,062} & -- \\
        \midrule
        \multirow{2}{*}{RL Set}
            & Each Edit Type & 704 & -- \\
            & \textit{Total} & \textit{3,520} & -- \\
        \midrule
        \multirow{6}{*}{VibeEdit Benchmark}
            & Addition & 96 & 18.97 \\
            & Removal & 80 & 19.66 \\
            & Attribute Modification & 83 & 20.57 \\
            & Object Replacement & 78 & 23.59 \\
            & Object Movement & 82 & 23.96 \\
            & \textit{Total} & \textit{419} & \textit{21.26} \\
        \bottomrule
    \end{tabular}
    \vspace{-4mm}
\end{table}

\section{Qualitative Results and Visualizations}
\label{sec:supp-visualizations}

\begin{figure}[ht]
\centering
\includegraphics[width=\textwidth]{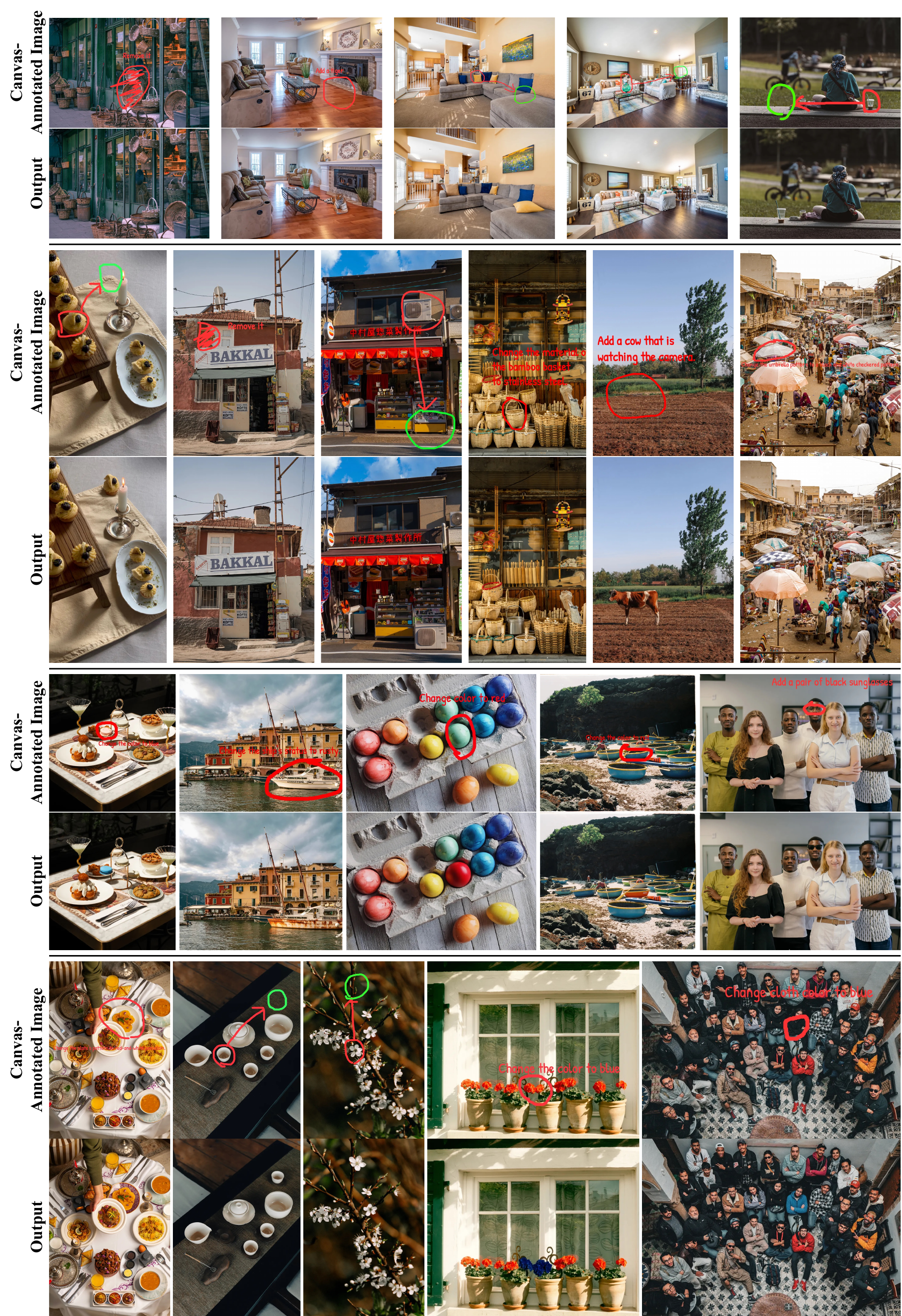}
\caption{Additional VibeEdit results across the five edit types. Each example
pairs a canvas-annotated image with the generated edit.}
\label{fig:more-demo}
\end{figure}

Figure~\ref{fig:more-demo} presents additional qualitative results across all
five edit types. These examples span diverse scenes, object scales, and
annotation forms, further illustrating that VibeEdit accurately interprets
canvas instructions, performs localized edits, and preserves unrelated
image content.

\clearpage
\section{System Prompts}
\label{sec:system-prompts}

This section summarizes the operative instructions used by model-based
components in our data and evaluation pipelines. We retain the rules needed to
understand and reproduce each stage, while omitting API syntax, exhaustive
output schemas, and retry logic.

\subsection{Object Selection Rules}
\label{sec:selection-prompts}

For each source image, Florence-2 proposals are assigned visible identifiers.
The selector receives the indexed image, optional candidate crops, and the
proposal list. It first applies shared hard rules and then the rules for the
requested operation. Addition is obtained by reversing a retained removal pair.

\newtcolorbox{rulebox}[1]{
  enhanced,
  colback=white,
  colframe=black!70,
  boxrule=0.8pt,
  arc=2pt,
  title={#1},
  colbacktitle=black!75,
  coltitle=white,
  fonttitle=\bfseries,
  fontupper=\scriptsize,
  left=8pt,right=8pt,top=5pt,bottom=5pt,
  before skip=4pt,after skip=4pt
}

\begin{rulebox}{Shared Hard Rules}
\raggedright
\begin{itemize}
  \setlength{\itemsep}{2pt}\setlength{\topsep}{1pt}
  \item Select only from the displayed proposals; never introduce an unboxed
  object.
  \item The target must be recognizable, sufficiently visible, mostly inside
  the frame, and spatially separable from its surroundings.
  \item The requested edit must be localized and visually verifiable. If no
  proposal supports the operation, reject the image.
\end{itemize}
\end{rulebox}

\begin{rulebox}{Addition and Removal}
\raggedright

\begin{itemize}
  \setlength{\itemsep}{2pt}\setlength{\topsep}{1pt}
  \item Prefer an isolated, medium-scale object with a clean boundary and a
  reconstructable background.
  \item Avoid entangled duplicates and large attached shadows or reflections.
\end{itemize}
\end{rulebox}

\begin{rulebox}{Attribute Modification}
\raggedright

\begin{itemize}
  \setlength{\itemsep}{2pt}\setlength{\topsep}{1pt}
  \item Choose an object with a clearly visible surface and edit exactly one
  family: color, material, texture, pattern, state, or object-local style.
  \item The source value must be observable and the target value distinct,
  plausible, and easy to verify.
  \item Preserve identity, geometry, pose, position, count, and non-target
  content; for people and animals, edit only external appearance.
\end{itemize}
\end{rulebox}

\begin{rulebox}{Object Replacement}
\raggedright

\begin{itemize}
  \setlength{\itemsep}{2pt}\setlength{\topsep}{1pt}
  \item Choose a well-bounded object with enough visible area for localized
  inpainting; avoid structural surfaces, text, reflections, and ambiguous
  regions.
  \item Use one concrete replacement from a different category with compatible
  context, visual style, and physical scale.
\end{itemize}
\end{rulebox}

\begin{rulebox}{Object Movement}
\raggedright

\begin{itemize}
  \setlength{\itemsep}{2pt}\setlength{\topsep}{1pt}
  \item Choose a movable, rigid object and a clearly separated empty
  destination on a physically compatible support.
  \item Reject ambiguous displacements, occupied destinations, and targets
  containing an existing similar object.
\end{itemize}
\end{rulebox}

\subsection{Per-Criterion Rubric Judge Prompt}
\label{sec:rubric-prompt}

We use a shared rubric and prompting protocol to assess edit
success, outside preservation, and local edit quality.
GPT-5.4-mini computes rewards during RL, while GPT-5.6-sol
serves as the benchmark evaluation judge. Each rubric criterion
is evaluated in a separate request containing the same visual
evidence and one binary question. The textual metadata below
describes the intended edit to the judge and is not provided
to the editing policy.

\begin{tcolorbox}[
  breakable,
  enhanced,
  colback=white,
  colframe=black!70,
  boxrule=0.8pt,
  arc=2pt,
  title={System Prompt},
  colbacktitle=black!75,
  coltitle=white,
  fonttitle=\bfseries,
  left=8pt,right=8pt,top=8pt,bottom=8pt
]
\scriptsize
\raggedright
\setlength{\parindent}{0pt}
\setlength{\parskip}{2pt}

You are an image editing evaluator. Judge localized image editing with
before/after evidence. Each request asks exactly one binary question about
either: (1) whether the intended edit region was edited correctly, (2) whether
regions outside the edit region stayed visually unchanged, (3) whether the
edited local region is visually well-formed and realistic, or (4) whether the
edited local region is aesthetically pleasing and naturally integrated.

The red rectangles and labels are evaluation overlays, not image content. Use
the original/source images to understand the pre-edit scene and the edited
images to judge the result. Use the full images for global context and the
side-by-side crops for precise local comparison. When
\texttt{object\_info} is provided, use it as the explicit summary of the
intended object or final object state.

Answer only valid JSON with keys \texttt{score} and \texttt{reasoning}.
\texttt{score} must be 1 only if the question-specific success criterion is
clearly satisfied. If the requested edit is near the red bbox but mainly
outside it, answer 0 for localization questions. For preservation questions,
ignore the intended edit bbox or bboxes and judge visible changes elsewhere in
the scene. For outside-unchanged questions, the bbox is an approximate edit
region, not an exact object mask: minor contiguous boundary spillover of the
intended edit is acceptable when the target object's main body is inside the
bbox. When uncertain, answer 0.
\end{tcolorbox}

\begin{tcolorbox}[
  breakable,
  enhanced,
  colback=white,
  colframe=black!70,
  boxrule=0.8pt,
  arc=2pt,
  title={User Prompt},
  colbacktitle=black!75,
  coltitle=white,
  fonttitle=\bfseries,
  left=8pt,right=8pt,top=8pt,bottom=8pt
]
\scriptsize
\raggedright
\setlength{\parindent}{0pt}
\setlength{\parskip}{2pt}

The bracketed fields below are populated at runtime; all other text is sent
verbatim.

\texttt{edit\_type: \{edit type\}}\newline
\texttt{target\_instruction: \{target instruction\}}\newline
\texttt{full\_instruction: \{full instruction\}}\newline
\texttt{object\_info (pre-parsed from the edit instruction; use this to
identify the target):}\newline
\hspace*{1em}\texttt{\{operation-specific object fields\}}\newline
\texttt{\{TARGET bbox line; movement uses SOURCE and TARGET bbox lines\}}

The source images show the scene before editing. The edited images show the
result after editing. Judge whether the requested change was correctly applied
in \texttt{\{edit region\}} and, for preservation questions, whether regions
outside \texttt{\{edit region\}} remain unchanged.

\texttt{\{boundary-tolerance block; outside-preservation criteria only\}}

\texttt{Image order:}\newline
\texttt{\{operation-specific image list\}}

\texttt{Question (\{rule name\}):}\newline
\texttt{\{one binary rubric question\}}

Return exactly one JSON object, for example:\newline
\texttt{\{"score": 0, "reasoning": "one short sentence"\}}
\end{tcolorbox}

The judge receives operation-specific object descriptions:
the object to add or remove, the original and target descriptions
for replacement or attribute modification, and the object name
for movement. Missing information is inferred from the instructions
and images. Each bounding box specifies its pixel coordinates,
size, fraction of the image area, and approximate location.
The edit region is defined by the TARGET box, or the union
of the SOURCE and TARGET boxes for movement.

For outside-preservation criteria, we allow minor changes
adjacent to the edit boundary, such as background cleanup after
removal and natural boundary effects. This tolerance does not
extend to misplaced objects, duplicates, or unrelated edits.
The visual inputs include annotated source and edited images
with aligned before/after crops of the edit region. Movement
examples include crops of both the source and destination
regions. Images precede the text in each request. The complete
rubric questions are listed in
Section~\ref{sec:complete-rubric-questions}.

\subsection{Complete Rubric Questions}
\label{sec:complete-rubric-questions}

The following boxes present the complete binary rubric used for
RL rewards and benchmark evaluation. Each example is assessed
using a common set of questions together with questions specific
to its edit type. The rubric covers three dimensions:
\emph{Edit Success}, which checks completion of the requested edit;
\emph{Outside Preservation}, which checks that unrelated content
remains unchanged; and \emph{Local Edit Quality}, which assesses
visual quality, scene integration, and aesthetics within the
edited region. Placeholders are filled with the corresponding
edit information before each judge call.

\newtcolorbox{rubricbox}[1]{
  breakable,
  enhanced,
  colback=white,
  colframe=black!70,
  boxrule=0.8pt,
  arc=2pt,
  title={#1},
  colbacktitle=black!75,
  coltitle=white,
  fonttitle=\bfseries,
  fontupper=\scriptsize,
  left=8pt,right=8pt,top=6pt,bottom=6pt,
  before skip=5pt,after skip=5pt
}

\begin{rubricbox}{Shared}
\raggedright
\textbf{Outside Preservation.}
\begin{enumerate}
  \setlength{\itemsep}{3pt}\setlength{\topsep}{2pt}
  \item \textbf{\texttt{vlm\_no\_global\_style\_drift}.}
  Compare the source full image with the edited full image. Outside the
  intended edit region or regions, does the image preserve the original global
  style, color palette, camera viewpoint, background layout, and overall scene
  identity? Answer 0 if the edited image has global color/style drift, changed
  viewpoint, changed background, changed scene layout, or looks like a
  regenerated different image.

  \item \textbf{\texttt{vlm\_instruction\_no\_extra\_edits}.}
  Did the edited image avoid unrelated extra edits beyond the requested visual
  instruction? Answer 1 only if the image changes only what is necessary for
  the requested edit. Answer 0 if unrelated objects are added, removed,
  replaced, moved, recolored, or noticeably altered.
\end{enumerate}

\textbf{Local Edit Quality.}
\begin{enumerate}
  \setlength{\itemsep}{3pt}\setlength{\topsep}{2pt}
  \item \textbf{\texttt{vlm\_local\_quality}.}
  Compare the source TARGET evidence, edited TARGET evidence, and edited full
  image. Is the edited TARGET region locally high-quality and visually
  well-formed? Answer 1 only if the edited object or filled region has coherent
  geometry, clean boundaries, plausible texture/detail, and no obvious
  deformation, blur, smear, broken structure, duplicated parts, or generation
  artifacts. Answer 0 if the local edit looks malformed, low-detail, melted,
  pasted, smeared, structurally broken, or technically poor, even if the
  requested edit is recognizable.

  \item \textbf{\texttt{vlm\_scene\_integration}.}
  Compare the source full image, edited full image, and edited TARGET evidence.
  Does the edited TARGET region fit naturally into the surrounding scene?
  Answer 1 only if scale, perspective, lighting direction, color temperature,
  shadows/contact, occlusion, and sharpness are plausible and consistent with
  nearby content. Answer 0 if the edit looks pasted on, floats, has wrong scale
  or perspective, mismatched lighting/color/sharpness, missing contact or
  shadow, or inconsistent occlusion with the scene.

  \item \textbf{\texttt{vlm\_local\_aesthetic}.}
  Compare the edited TARGET evidence and edited full image. Is the edited
  TARGET region locally aesthetically pleasing while preserving realism?
  Answer 1 only if the edit looks natural, visually harmonious, and not
  distracting or ugly in the local composition, with appealing shape,
  material, color, and detail for this scene. Answer 0 if the edit is
  semantically correct but visually unpleasant, awkward, cheap-looking,
  cluttered, jarring, or aesthetically worse than a plausible natural edit.
\end{enumerate}
\end{rubricbox}

\begin{rubricbox}{Object Addition}
\raggedright
\textbf{Edit Success.}
\begin{enumerate}
  \setlength{\itemsep}{3pt}\setlength{\topsep}{2pt}
  \item \textbf{\texttt{vlm\_add\_correct\_in\_target}.}
  Compare the original/source TARGET evidence with the edited TARGET evidence.
  Was \texttt{\{object\_desc\}} added clearly and primarily inside the red
  TARGET bbox, while fitting the local scene? Answer 1 only if
  \texttt{\{object\_desc\}} is recognizable in the TARGET bbox and not mainly
  outside it.
\end{enumerate}

\textbf{Outside Preservation.}
\begin{enumerate}
  \setlength{\itemsep}{3pt}\setlength{\topsep}{2pt}
  \item \textbf{\texttt{vlm\_add\_outside\_unchanged}.}
  Compare the original/source full image with the edited full image. Outside
  the red TARGET bbox, do the scene, objects, colors, lighting, and layout
  remain visually unchanged? Use the boundary tolerance above for
  \texttt{\{object\_desc\}}. Answer 0 if unrelated regions, background, or
  global appearance changed, if an extra copy of \texttt{\{object\_desc\}}
  appears outside the TARGET bbox, or if the main body of
  \texttt{\{object\_desc\}} is outside the TARGET bbox and only part of it
  extends into the bbox.

  \item \textbf{\texttt{vlm\_add\_no\_duplicate\_elsewhere}.}
  Compare the edited full image with the source full image. Was
  \texttt{\{object\_desc\}} added only once, without an extra duplicate or
  copy appearing elsewhere outside the TARGET bbox? Answer 0 if another
  instance of \texttt{\{object\_desc\}} appears outside the TARGET bbox due to
  the edit.
\end{enumerate}

\textbf{Local Edit Quality.}
\begin{enumerate}
  \setlength{\itemsep}{3pt}\setlength{\topsep}{2pt}
  \item \textbf{\texttt{vlm\_add\_contact\_occlusion}.}
  Does the added \texttt{\{object\_desc\}} have plausible contact, support,
  shadow, and occlusion with the surrounding scene? Answer 0 if it floats,
  cuts through other objects incorrectly, lacks expected contact or shadow, or
  appears pasted on.
\end{enumerate}
\end{rubricbox}

\begin{rubricbox}{Object Removal}
\raggedright
\textbf{Edit Success.}
\begin{enumerate}
  \setlength{\itemsep}{3pt}\setlength{\topsep}{2pt}
  \item \textbf{\texttt{vlm\_remove\_absent\_in\_target}.}
  Compare the original/source TARGET evidence with the edited TARGET evidence.
  Is \texttt{\{object\_desc\}}, the original target object, absent from the
  edited red TARGET bbox? Answer 1 only if \texttt{\{object\_desc\}} is no
  longer visible in the edited TARGET bbox.

  \item \textbf{\texttt{vlm\_remove\_natural\_fill}.}
  After removing \texttt{\{object\_desc\}}, does the edited red TARGET bbox
  look naturally filled, without obvious holes, smears, duplicated texture, or
  broken structure?

  \item \textbf{\texttt{vlm\_remove\_no\_remnant}.}
  Compare the source TARGET evidence with the edited TARGET evidence. Are there
  no visible remnants, ghost traces, outlines, partial parts, shadows, or
  recognizable fragments of \texttt{\{object\_desc\}}? Answer 0 if any clear
  remnant of the removed object remains.
\end{enumerate}

\textbf{Outside Preservation.}
\begin{enumerate}
  \setlength{\itemsep}{3pt}\setlength{\topsep}{2pt}
  \item \textbf{\texttt{vlm\_remove\_outside\_unchanged}.}
  Compare the original/source full image with the edited full image. Outside
  the red TARGET bbox, do the scene, objects, colors, lighting, and layout
  remain visually unchanged? Ignore small natural cleanup or fill changes that
  touch the TARGET bbox boundary and are directly caused by removing
  \texttt{\{object\_desc\}}. Answer 0 if unrelated regions, objects,
  background, or global appearance changed, if \texttt{\{object\_desc\}}
  appears shifted or copied outside the TARGET bbox, or if the main body of
  \texttt{\{object\_desc\}} remains outside the bbox as an unremoved object.
\end{enumerate}

\textbf{Local Edit Quality.}
\begin{enumerate}
  \setlength{\itemsep}{3pt}\setlength{\topsep}{2pt}
  \item \textbf{\texttt{vlm\_remove\_background\_continuity}.}
  Does the filled TARGET region continue the surrounding background, texture,
  geometry, lighting, and perspective naturally? Answer 0 if the removed area
  has obvious repeated texture, smear, hole, blur patch, warped lines, or
  inconsistent background structure.
\end{enumerate}
\end{rubricbox}

\begin{rubricbox}{Object Replacement}
\raggedright
\textbf{Edit Success.}
\begin{enumerate}
  \setlength{\itemsep}{3pt}\setlength{\topsep}{2pt}
  \item \textbf{\texttt{vlm\_replace\_correct\_in\_target}.}
  Compare the original/source TARGET evidence with the edited TARGET evidence.
  Was \texttt{\{original\_object\_desc\}} replaced by
  \texttt{\{object\_desc\}} clearly and primarily inside the red TARGET bbox?
  Answer 1 only if \texttt{\{original\_object\_desc\}} is no longer the
  visible target and \texttt{\{object\_desc\}} is recognizable in the TARGET
  bbox.

  \item \textbf{\texttt{vlm\_replace\_old\_object\_absent}.}
  Compare the source TARGET evidence with the edited TARGET evidence. Is
  \texttt{\{original\_object\_desc\}} no longer visible as the main object in
  the TARGET bbox? Answer 0 if the original object remains, is only partially
  changed, or is still recognizable as the target.

  \item \textbf{\texttt{vlm\_replace\_new\_object\_identity\_clear}.}
  Is the replacement object clearly recognizable as
  \texttt{\{object\_desc\}}, not an ambiguous hybrid or malformed mixture of
  the old and new objects? Answer 0 if the result is a hybrid, unclear object,
  or does not clearly match the requested replacement.
\end{enumerate}

\textbf{Outside Preservation.}
\begin{enumerate}
  \setlength{\itemsep}{3pt}\setlength{\topsep}{2pt}
  \item \textbf{\texttt{vlm\_replace\_outside\_unchanged}.}
  Compare the original/source full image with the edited full image. Outside
  the red TARGET bbox, do the scene, objects, colors, lighting, and layout
  remain visually unchanged? Use the boundary tolerance above for
  \texttt{\{object\_desc\}}. Answer 0 if unrelated regions changed, if an
  extra copy of \texttt{\{object\_desc\}} appears outside the TARGET bbox, or
  if the main body of \texttt{\{object\_desc\}} is outside the TARGET bbox and
  only part of it extends into the bbox.
\end{enumerate}

\end{rubricbox}

\begin{rubricbox}{Object Attribute Modification}
\raggedright
\textbf{Edit Success.}
\begin{enumerate}
  \setlength{\itemsep}{3pt}\setlength{\topsep}{2pt}
  \item \textbf{\texttt{vlm\_attribute\_correct\_in\_target}.}
  Compare the original/source TARGET evidence with the edited TARGET evidence.
  Does the target object inside the red TARGET bbox clearly match
  \texttt{\{object\_desc\}}, while remaining plausible and recognizable?
  Answer 1 only if the requested edit is visible on the TARGET object, not
  only on a nearby or outside object.

  \item \textbf{\texttt{vlm\_attribute\_same\_object\_identity}.}
  Compare the original/source TARGET evidence with the edited TARGET evidence.
  Is the edited TARGET still the same object or object instance as
  \texttt{\{original\_object\_desc\}}, with only the requested attribute
  modified to match \texttt{\{object\_desc\}}? Answer 0 if the target was
  replaced by a different object, changed category, lost its main
  shape/structure, or gained unrelated edits beyond the requested attribute.

  \item \textbf{\texttt{vlm\_attribute\_only\_requested\_attribute\_changed}.}
  Compare the source TARGET evidence with the edited TARGET evidence. Was only
  the requested attribute modified, while the object's category, identity,
  pose, shape, size, and location remained essentially the same? Answer 0 if
  the object was replaced, moved, reshaped substantially, changed category, or
  gained unrelated changes.
\end{enumerate}

\textbf{Outside Preservation.}
\begin{enumerate}
  \setlength{\itemsep}{3pt}\setlength{\topsep}{2pt}
  \item \textbf{\texttt{vlm\_attribute\_outside\_unchanged}.}
  Compare the original/source full image with the edited full image. Outside
  the red TARGET bbox, do the scene, objects, colors, lighting, and layout
  remain visually unchanged? Use the boundary tolerance above for the edited
  target object. Answer 0 if unrelated regions changed, if the requested
  edit toward \texttt{\{object\_desc\}} was applied to another object
  outside the TARGET bbox, or if the changed object's main body is outside the
  TARGET bbox and only part of it extends into the bbox.

  \item \textbf{\texttt{vlm\_attribute\_no\_spillover}.}
  Was the requested edit confined to the target object, without changing
  nearby objects or background regions? Answer 0 if the same edit
  appears on non-target objects or surrounding background.
\end{enumerate}

\end{rubricbox}

\begin{rubricbox}{Object Movement}
\raggedright
\textbf{Edit Success.}
\begin{enumerate}
  \setlength{\itemsep}{3pt}\setlength{\topsep}{2pt}
  \item \textbf{\texttt{vlm\_move\_source\_cleared}.}
  Is the \texttt{\{object\_name\}} no longer visible inside the red SOURCE
  bbox? Answer 1 only if the SOURCE bbox is cleared of the
  \texttt{\{object\_name\}}.

  \item \textbf{\texttt{vlm\_move\_target\_contains\_object}.}
  Compare the original/source SOURCE evidence with the edited TARGET evidence.
  Is the \texttt{\{object\_name\}} clearly visible and primarily located
  inside the red TARGET bbox? Answer 1 only if the TARGET bbox contains the
  \texttt{\{object\_name\}} requested by the instruction.

  \item \textbf{\texttt{vlm\_move\_no\_duplicate}.}
  Compare the original/source and edited full images plus both SOURCE and
  TARGET crops. Is there no obvious duplicate of the
  \texttt{\{object\_name\}} left elsewhere in the edited image? Answer 1 only
  if the \texttt{\{object\_name\}} appears moved from SOURCE to TARGET without
  leaving a clear duplicate copy.

  \item \textbf{\texttt{vlm\_move\_object\_identity\_preserved}.}
  Does the moved object keep the same identity, category, and recognizable
  appearance as the original source \texttt{\{object\_name\}}? Answer 0 if the
  moved object changes into a different object, loses key identity, or becomes
  unrecognizable.
\end{enumerate}

\textbf{Outside Preservation.}
\begin{enumerate}
  \setlength{\itemsep}{3pt}\setlength{\topsep}{2pt}
  \item \textbf{\texttt{vlm\_move\_outside\_unchanged}.}
  Compare the original/source full image with the edited full image. Outside
  the union of the red SOURCE and TARGET bboxes, do the scene, objects, colors,
  lighting, and layout remain visually unchanged? Use the boundary tolerance
  above for the moved \texttt{\{object\_name\}}. Answer 0 if unrelated
  regions, background, or global appearance changed, if the main body of the
  moved \texttt{\{object\_name\}} is outside the TARGET bbox and only part of
  it extends into the bbox, or if an obvious duplicate of the
  \texttt{\{object\_name\}} remains outside the SOURCE/TARGET bbox union.
\end{enumerate}

\textbf{Local Edit Quality.}
\begin{enumerate}
  \setlength{\itemsep}{3pt}\setlength{\topsep}{2pt}
  \item \textbf{\texttt{vlm\_move\_source\_natural\_fill}.}
  After \texttt{\{object\_name\}} was moved away, does the SOURCE bbox look
  naturally filled or restored without obvious holes, smears, ghost traces, or
  broken background? Answer 0 if the source area contains remnants, shadows,
  holes, smeared texture, or unnatural fill artifacts.
\end{enumerate}
\end{rubricbox}

\end{document}